\documentclass[5p]{elsarticle}

\usepackage{lineno,hyperref}

\usepackage{soul}
\soulregister\cite7 
\soulregister\citep7 
\soulregister\citet7 
\soulregister\ref7 
\soulregister\pageref7 
\usepackage{mathrsfs}
\usepackage[noend]{algpseudocode}

\usepackage{algorithmicx,algorithm}
\usepackage{bbm}
\usepackage{graphicx}
\usepackage{amsmath}
\usepackage{amssymb}
\usepackage{multirow}
\usepackage{helvet}
\usepackage{courier}
\usepackage{float}

\usepackage{amsfonts}
\usepackage{textcomp}
\usepackage{hyperref}

\usepackage{hyperref}
\usepackage{times}
\usepackage{epsfig}
\usepackage[usenames,dvipsnames]{color}

\usepackage{amsmath,amsfonts}

\usepackage{textcomp}
\usepackage{stfloats}
\usepackage{url}
\usepackage{verbatim}
\usepackage{arydshln}
\usepackage{bbding}
\usepackage{wrapfig}
\usepackage{caption}
\usepackage{booktabs}
\usepackage{newfloat}
\usepackage{listings}
\usepackage{bm}

\usepackage{stackengine}
\usepackage[table,xcdraw]{xcolor} 
\usepackage{colortbl} 
\usepackage{subcaption}

\usepackage{pifont}
\usepackage{bbding}

\usepackage{makecell}

\definecolor{ceiling}{RGB}{210,42,33}
\definecolor{floor}{RGB}{40,159,7}
\definecolor{wall}{RGB}{162,215,224}
\definecolor{windows}{RGB}{113,158,201}
\definecolor{chair}{RGB}{202,205,76}
\definecolor{bed}{RGB}{224,190,159}
\definecolor{sofa}{RGB}{153,98,192}
\definecolor{table}{RGB}{23,124,181}
\definecolor{tvs}{RGB}{160,187,34}
\definecolor{furniture}{RGB}{225,126,18}
\definecolor{objects}{RGB}{199,174,221}

\definecolor{road}{RGB}{255,1,252}
\definecolor{sidewalk}{RGB}{76,0,74}
\definecolor{parking}{RGB}{255,149,255}
\definecolor{othergrnd}{RGB}{178,4,75}
\definecolor{building}{RGB}{255,198,0}
\definecolor{car}{RGB}{98,152,240}
\definecolor{truck}{RGB}{82,28,183}
\definecolor{bicycle}{RGB}{101,229,248}
\definecolor{motorcycle}{RGB}{35,57,148}
\definecolor{otherveh}{RGB}{100,82,245}
\definecolor{vegetation}{RGB}{2,176,0}
\definecolor{trunk}{RGB}{131,62,6}
\definecolor{terrain}{RGB}{153,238,85}
\definecolor{person}{RGB}{255,26,29}
\definecolor{bicyclist}{RGB}{255,40,202}
\definecolor{motorcycl}{RGB}{150,29,99}
\definecolor{fence}{RGB}{148,125,42}
\definecolor{pole}{RGB}{255,239,143}
\definecolor{trafsign}{RGB}{248,2,3}

\makeatletter
\newcommand{\car@semkitfreq}{3.92}
\newcommand{\bicycle@semkitfreq}{0.03}
\newcommand{\motorcycle@semkitfreq}{0.03}
\newcommand{\truck@semkitfreq}{0.16}
\newcommand{\othervehicle@semkitfreq}{0.20}
\newcommand{\person@semkitfreq}{0.07}
\newcommand{\bicyclist@semkitfreq}{0.07}
\newcommand{\motorcyclist@semkitfreq}{0.05}
\newcommand{\road@semkitfreq}{15.30}  %
\newcommand{\parking@semkitfreq}{1.12}
\newcommand{\sidewalk@semkitfreq}{11.13}  %
\newcommand{\otherground@semkitfreq}{0.56}
\newcommand{\building@semkitfreq}{14.1}  %
\newcommand{\fence@semkitfreq}{3.90}
\newcommand{\vegetation@semkitfreq}{39.3}  %
\newcommand{\trunk@semkitfreq}{0.51}
\newcommand{\terrain@semkitfreq}{9.17} %
\newcommand{\pole@semkitfreq}{0.29}
\newcommand{\trafficsign@semkitfreq}{0.08}
\newcommand{\semkitfreq}[1]{{\csname #1@semkitfreq\endcsname}}

\definecolor{nbarrier}{RGB}{112, 128, 144}
\definecolor{nbicycle}{RGB}{220, 20, 60}
\definecolor{nbus}{RGB}{255, 127, 80}
\definecolor{ncar}{RGB}{255, 158, 0}
\definecolor{nconstruct}{RGB}{233, 150, 70}
\definecolor{nmotor}{RGB}{255, 61, 99}
\definecolor{npedestrian}{RGB}{0, 0, 230}
\definecolor{ntraffic}{RGB}{47, 79, 79}
\definecolor{ntrailer}{RGB}{255, 140, 0}
\definecolor{ntruck}{RGB}{255, 99, 71}
\definecolor{ndriveable}{RGB}{0, 207, 191}
\definecolor{nother}{RGB}{175, 0, 75}
\definecolor{nsidewalk}{RGB}{75, 0, 75}
\definecolor{nterrain}{RGB}{112, 180, 60}
\definecolor{nmanmade}{RGB}{222, 184, 135}
\definecolor{nvegetation}{RGB}{0, 175, 0}
\definecolor{nothers}{RGB}{0, 0, 0}

\newcommand{\car@sscbkitfreq}{2.85}
\newcommand{\bicycle@sscbkitfreq}{0.01}
\newcommand{\motorcycle@sscbkitfreq}{0.01}
\newcommand{\truck@sscbkitfreq}{0.16}
\newcommand{\otherveh@sscbkitfreq}{5.75}
\newcommand{\person@sscbkitfreq}{0.02}
\newcommand{\road@sscbkitfreq}{14.98}
\newcommand{\parking@sscbkitfreq}{2.31}
\newcommand{\sidewalk@sscbkitfreq}{6.43}
\newcommand{\othergrnd@sscbkitfreq}{2.05}
\newcommand{\building@sscbkitfreq}{15.67}
\newcommand{\fence@sscbkitfreq}{0.96}
\newcommand{\vegetation@sscbkitfreq}{41.99}
\newcommand{\terrain@sscbkitfreq}{7.10}
\newcommand{\pole@sscbkitfreq}{0.22}
\newcommand{\trafsign@sscbkitfreq}{0.06}
\newcommand{\otherstructure@sscbkitfreq}{4.33}
\newcommand{\otherobject@sscbkitfreq}{0.28}
\newcommand{\sscbkitfreq}[1]{{\csname #1@sscbkitfreq\endcsname}}

\definecolor{otherstructure}{rgb}{0.98039215, 0.58823529, 0.}
\definecolor{otherobject}{rgb}{0.19607843, 1.        , 1.        }

\def \cmk {\checkmark}
\def \xmk {\ding{55}}

\begin{document}

\begin{sloppypar}
\begin{frontmatter}

\title{A Survey on Occupancy Perception for Autonomous Driving: The Information Fusion Perspective}

\author{Huaiyuan Xu}
\author{Junliang Chen}
\author{Shiyu Meng}
\author{Yi Wang}
\author{Lap-Pui Chau}

\address{Department of Electrical and Electronic Engineering, The Hong Kong Polytechnic University, Hong Kong SAR}
\fntext[myfootnote]{Lap-Pui Chau is the corresponding author.}

\date{} 

\begin{abstract}
3D occupancy perception technology aims to observe and understand dense 3D environments for autonomous vehicles. Owing to its comprehensive perception capability, this technology is emerging as a trend in autonomous driving perception systems, and is attracting significant attention from both industry and academia. Similar to traditional bird's-eye view (BEV) perception, 3D occupancy perception has the nature of multi-source input and the necessity for information fusion. However, the difference is that it captures vertical structures that are ignored by 2D BEV. In this survey, we review the most recent works on 3D occupancy perception, and provide in-depth analyses of methodologies with various input modalities. Specifically, we summarize general network pipelines, highlight information fusion techniques, and discuss effective network training. We evaluate and analyze the occupancy perception performance of the state-of-the-art on the most popular datasets. Furthermore, challenges and future research directions are discussed. We hope this paper will inspire the community and encourage more research work on 3D occupancy perception. A comprehensive list of studies in this survey is publicly available in an active repository that continuously collects the latest work: \href{https://github.com/HuaiyuanXu/3D-Occupancy-Perception}{https://github.com/HuaiyuanXu/3D-Occupancy-Perception}.

\end{abstract}

\end{frontmatter}


\section{Introduction}
\label{sec:intro}
\subsection{Occupancy Perception in Autonomous Driving}
Autonomous driving can improve urban traffic efficiency and reduce energy consumption. For reliable and safe autonomous driving, a crucial capability is to understand the surrounding environment, that is, to perceive the observed world. At present, bird's-eye view (BEV) perception is the mainstream perception pattern \cite{li2023delving,ma2022vision}, with the advantages of absolute scale and no occlusion for describing environments. BEV perception provides a unified representation space for multi-source information fusion (\textit{e.g.}, information from diverse viewpoints, modalities, sensors, and time series) and numerous downstream applications (\textit{e.g.}, explainable decision making and motion planning). However, BEV perception does not monitor height information, thereby cannot provide a complete representation for the 3D scene.

To address this, occupancy perception was proposed for autonomous driving to capture the dense 3D structure of the real world. This burgeoning perception technology aims to infer the occupied state of each voxel for the voxelized world, characterized by a strong generalization capability to open-set objects, irregular-shaped vehicles, and special road structures \cite{occupancy_network_blog,wang2023panoocc}. Compared with 2D views such as perspective view and bird's-eye view, occupancy perception has a nature of 3D attributes, making it more suitable for 3D downstream tasks, such as 3D detection \cite{peng2023learning,zhou2024sogdet}, segmentation \cite{wang2023panoocc}, and tracking \cite{min2024driveworld}.

\begin{figure}[t]
\setlength{\belowcaptionskip}{0pt}
    \centering
    \includegraphics[scale=0.29]{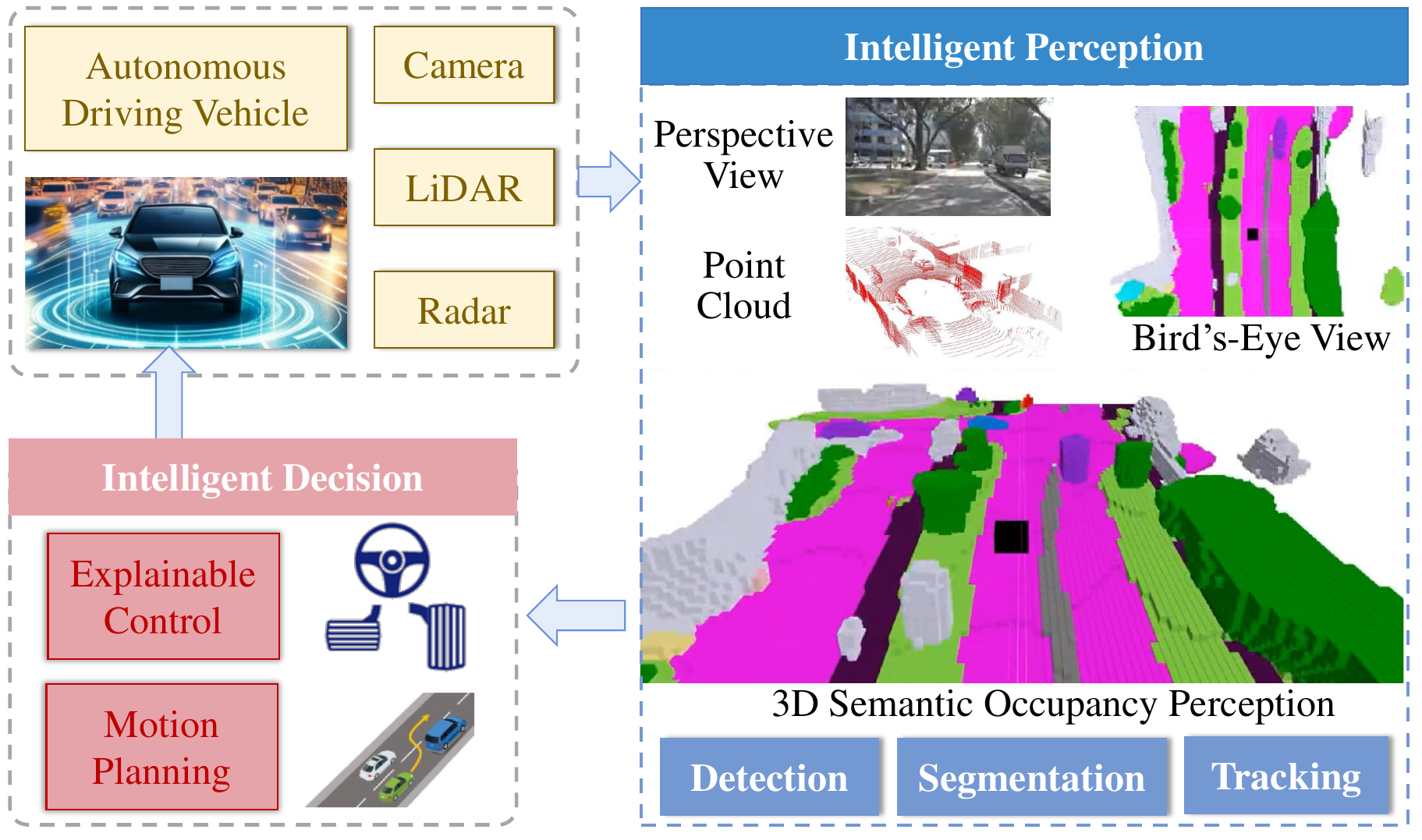}
    \caption{\textbf{Autonomous driving vehicle system.} The sensing data from cameras, LiDAR, and radar enable the vehicle to intelligently perceive its surroundings. Subsequently, the intelligent decision module generates control and planning of driving behavior. Occupancy perception surpasses other perception methods based on perspective view, bird's-eye view, or point clouds, in terms of 3D understanding and density.}
    \label{fig:big pic and illustration}
\end{figure}

In academia and industry, occupancy perception for holistic 3D scene understanding poses a meaningful impact. On the academic consideration, it is challenging to estimate dense 3D occupancy of the real world from complex input formats, encompassing multiple sensors, modalities, and temporal sequences. Moreover, it is valuable to further reason about semantic categories \cite{openocc}, textual descriptions \cite{vobecky2024pop}, and motion states \cite{cam4docc} for occupied voxels, which paves the way toward a more comprehensive understanding of the environment. From the industrial perspective, the deployment of a LiDAR Kit on each autonomous vehicle is expensive. With cameras as a cheap alternative to LiDAR, vision-centric occupancy perception is indeed a cost-effective solution that reduces the manufacturing cost for vehicle equipment manufacturers.

\subsection{Motivation to Information Fusion Research}

The gist of occupancy perception lies in comprehending complete and dense 3D scenes, including understanding occluded areas. However, the observation from a single sensor only captures parts of the scene. For instance, Fig. \ref{fig:big pic and illustration} intuitively illustrates that an image or a point cloud cannot provide a 3D panorama or a dense environmental scan. To this end, studying information fusion from multiple sensors \cite{openoccupancy,ming2024occfusion,song2024collaborative} and multiple frames \cite{wang2023panoocc,openocc} will facilitate a more comprehensive perception. This is because, on the one hand, information fusion expands the spatial range of perception, and on the other hand, it densifies scene observation. Besides, for occluded regions, integrating multi-frame observations is beneficial, as the same scene is observed by a host of viewpoints, which offer sufficient scene features for occlusion inference.

Furthermore, in complex outdoor scenarios with varying lighting and weather conditions, the need for stable occupancy perception is paramount. This stability is crucial for ensuring driving safety. At this point, research on multi-modal fusion will promote robust occupancy perception, by combining the strengths of different modalities of data \cite{openoccupancy,ming2024occfusion,wolters2024unleashing,sze2024real}. For example, LiDAR and radar data are insensitive to illumination changes and can sense the precise depth of the scene. This capability is particularly important during nighttime driving or in scenarios where the shadow/glare obscures critical information. Camera data excel in capturing detailed visual texture, being adept at identifying color-based environmental elements (\textit{e.g.}, road signs and traffic lights) and long-distance objects. Therefore, the fusion of data from LiDAR, radar, and camera will present a holistic understanding of the environment meanwhile against adverse environmental changes. 

\begin{figure*}[t]
\setlength{\belowcaptionskip}{0pt}
  \centering  
  \includegraphics[scale=0.283]{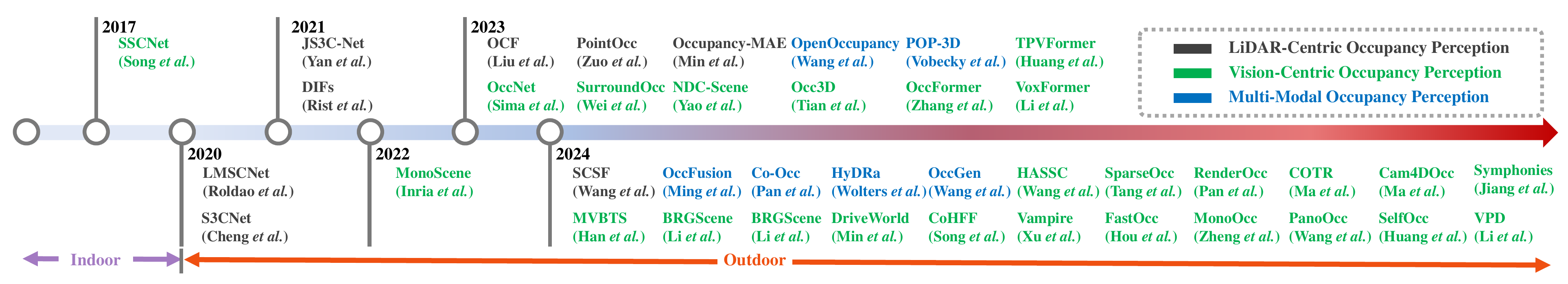}
    \caption{\textbf{Chronological overview of 3D occupancy perception.} It can be observed that: (1) research on occupancy has undergone explosive growth since 2023; (2) the predominant trend focuses on vision-centric occupancy, supplemented by LiDAR-centric and multi-modal methods.}
    \label{fig:Chronological overview}
\end{figure*}

\subsection{Contributions}
Among perception-related topics, 3D semantic segmentation \cite{xie2020linking,zhang2019review} and 3D object detection \cite{ma20233d,mao20233d,wang2023multi,fernandes2021point} have been extensively reviewed. However, these tasks do not facilitate a dense understanding of the environment. BEV perception, which addresses this issue, has also been thoroughly reviewed \cite{li2023delving,ma2022vision}. Our survey focuses on 3D occupancy perception, which captures the environmental height information overlooked by BEV perception. There are two related reviews: Roldao \textit{et al.} \cite{roldao20223d} conducted a literature review on 3D scene completion for both indoor and outdoor scenes; Zhang \textit{et al.} \cite{zhang2024vision} only reviewed 3D occupancy prediction based on the visual modality. Unlike their work, our survey is tailored to autonomous driving scenarios, and extends the existing 3D occupancy survey by considering more sensor modalities. Moreover, given the multi-source nature of 3D occupancy perception, we provide an in-depth analysis of information fusion techniques for this field. The primary contributions of this survey are three-fold:
\begin{itemize}
    \item We systematically review the latest research on 3D occupancy perception in the field of autonomous driving, covering motivation analysis, the overall research background, and an in-depth discussion on methodology, evaluation, and challenges.    
    \item We provide a taxonomy of 3D occupancy perception, and elaborate on core methodological issues, including network pipelines, multi-source information fusion, and effective network training.
    \item We present evaluations for 3D occupancy perception, and offer detailed performance comparisons. Furthermore, current limitations and future research directions are discussed.
\end{itemize}
The remainder of this paper is structured as follows. Sec. \ref{sec:Background} provides a brief background on the history, definitions, and related research domains. Sec. \ref{sec:Methodologies} details methodological insights. Sec. \ref{sec:Evaluation} conducts performance comparisons and analyses. Finally, future research directions are discussed and the survey is concluded in Sec. \ref{sec:Challenges and Opportunities} and \ref{sec:Conclusions}, respectively.

\section{Background}
\label{sec:Background}

\subsection{A Brief History of Occupancy Perception}
Occupancy perception is derived from Occupancy Grid Mapping (OGM) \cite{thrun2002probabilistic}, which is a classic topic in mobile robot navigation, and aims to generate a grid map from noisy and uncertain measurements. Each grid in this map is assigned a value that scores the probability of the grid space being occupied by obstacles. Semantic occupancy perception originates from SSCNet \cite{song2017semantic}, which predicts the occupied status and semantics of all voxels in an indoor scene from a single image. However, studying occupancy perception in outdoor scenes is imperative for autonomous driving, as opposed to indoor scenes. MonoScene \cite{monoscene} is a pioneering work of outdoor scene occupancy perception using only a monocular camera. Contemporary with MonoScene, Tesla announced its brand-new camera-only occupancy network at the CVPR 2022 workshop on Autonomous Driving \cite{WAD_2022}. This new network comprehensively understands the 3D environment surrounding a vehicle according to surround-view RGB images. Subsequently, occupancy perception has attracted extensive attention, catalyzing a surge in research on occupancy perception for autonomous driving in recent years. The chronological overview in Fig. \ref{fig:Chronological overview} indicates rapid development in occupancy perception since 2023. 

Early approaches to outdoor occupancy perception primarily used LiDAR input to infer 3D occupancy \cite{roldao2020lmscnet,yan2021sparse,cheng2021s3cnet}. However, recent methods have shifted towards more challenging vision-centric 3D occupancy prediction \cite{gan2023simple,huang2023tri,li2023voxformer,li2024one}. 
Presently, a dominant trend in occupancy perception research is vision-centric solutions, complemented by LiDAR-centric methods and multi-modal approaches. Occupancy perception can serve as a unified representation of the 3D physical world within the end-to-end autonomous driving framework \cite{openocc,hu2023planning}, followed by downstream applications spanning various driving tasks such as detection, tracking, and planning. The training of occupancy perception networks heavily relies on dense 3D occupancy labels, leading to the development of diverse street view occupancy datasets \cite{openoccupancy,cam4docc,occ3d,sscbench}. Recently, taking advantage of the powerful performance of large models, the integration of large models with occupancy perception has shown promise in alleviating the need for cumbersome 3D occupancy labeling \cite{zhang2023occnerf}.

\subsection{Task Definition}
\begin{figure}[t]
\setlength{\belowcaptionskip}{0pt}
    \centering
    \includegraphics[scale=0.29]{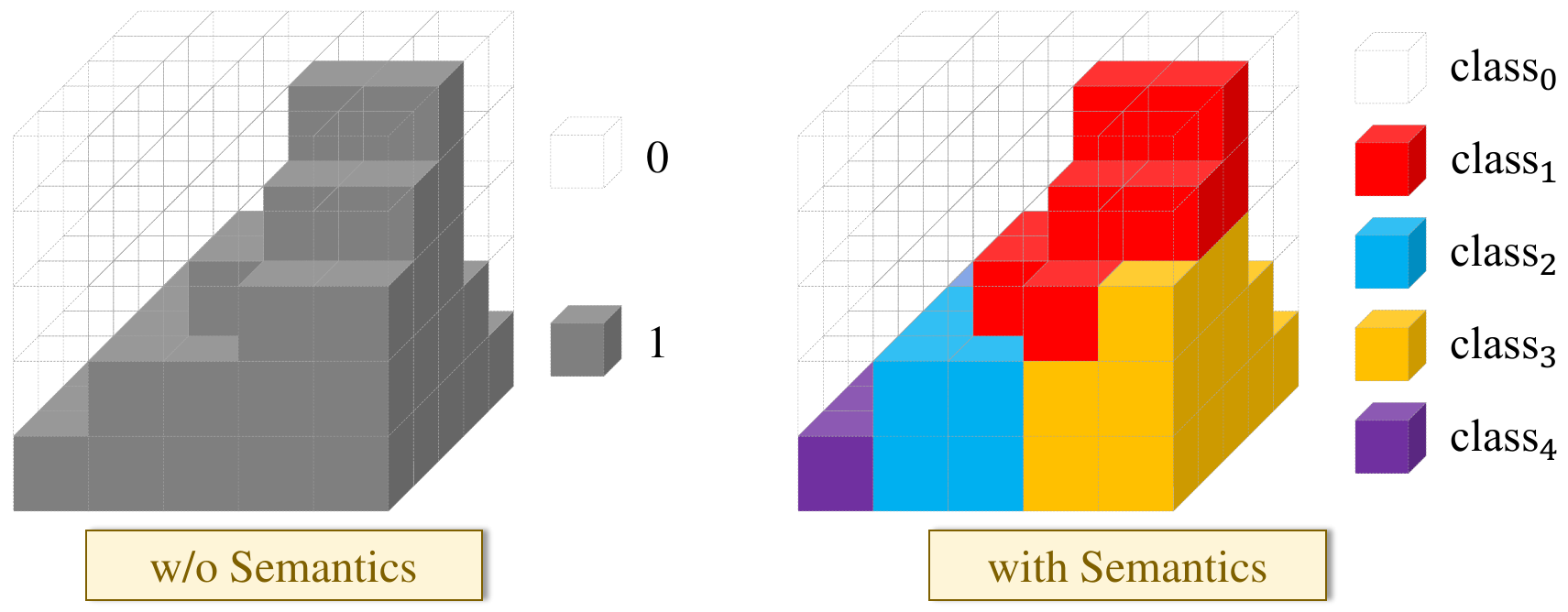}
    \caption{\textbf{Illustration of voxel-wise representations with and without semantics.} The left voxel volume depicts the overall occupancy distribution. The right voxel volume incorporates semantic enrichment, where each voxel is associated with a class estimation.}
    \label{fig:voxel-wise representation}
\end{figure}

Occupancy perception aims to extract voxel-wise representations of observed 3D scenes from multi-source inputs. Specifically, this representation involves discretizing a continuous 3D space $W$ into a grid volume $V$ composed of dense voxels. The state of each voxel is described by the value of $\left\{ 1,0\right\}$ or $\left\{ c_{0},\cdots,c_{n} \right\}$, as illustrated in Fig. \ref{fig:voxel-wise representation},
\begin{equation}
    W\in \mathbb{R}^{3}\to V\in\left\{ 0,1\right\}^{X\times Y\times Z} \mathrm{or} \left\{ c_{0},\cdots,c_{n} \right\}^{X\times Y\times Z},
\label{eq:voxel-wise representation}
\end{equation}
where $0$ and $1$ denote the occupied state; $c$ represents semantics; $\left ( X, Y, Z \right ) $ are the length, width, and height of the voxel volume. This voxelized representation offers two primary advantages: (1) it enables the transformation of unstructured data into a voxel volume, thereby facilitating processing by convolution \cite{he2016deep} and Transformer \cite{vaswani2017attention} architectures; (2) it provides a flexible and scalable representation for 3D scene understanding, striking an optimal trade-off between spatial granularity and memory consumption.

Multi-source input encompasses signals from multiple sensors, modalities, and frames, including common formats such as images and point clouds. We take the multi-camera images $\left \{ I_{t}^{1},\dots, I_{t}^{N}  \right \} $ and point cloud $P_{t}$ of the $t$-th frame as an input $\Omega_{t}=\left \{ I_{t}^{1},\dots, I_{t}^{N}, P_{t} \right \} $. $N$ is the number of cameras. The occupancy perception network $\Phi_{O}$ processes information from the $t$-th frame and the previous $k$ frames, generating the voxel-wise representation $V_{t}$ of the $t$-th frame:
\begin{equation}
    V_{t}=\Phi _{O}\left ( \Omega_{t},\dots,\Omega_{t-k} \right ) , \quad\mathrm{s.t.} \quad t-k\ge 0.
\label{eq:occupancy perception network}
\end{equation}

\subsection{Related Works}
\subsubsection{Bird's-Eye-View Perception}
Bird's-eye-view perception represents the 3D scene on a BEV plane. Specifically, it extracts the feature of each entire pillar in 3D space as the feature of the corresponding BEV grid. This compact representation provides a clear and intuitive depiction of the spatial layout from a top-down perspective. Tesla released its BEV perception-based systematic pipeline \cite{TeslaAIDay2021}, which is capable of detecting objects and lane lines in BEV space, for Level $2$ highway navigation and smart summoning.

According to the input data, BEV perception is primarily categorized into three groups: BEV camera \cite{zhang2022beverse,huang2021bevdet,li2022bevformer}, BEV LiDAR \cite{yang2018pixor,yang2018hdnet}, and BEV fusion \cite{liu2023bevfusion,liang2022bevfusion}. Current research predominantly focuses on the BEV camera, the key of which lies in the effective feature conversion from image space to BEV space. To address this challenge, one type of work adopts explicit transformation, which initially estimates the depth for front-view images, then utilizes the camera's intrinsic and extrinsic matrices to map image features into 3D space, and subsequently engages in BEV pooling \cite{huang2021bevdet,liang2022bevfusion,li2023bevdepth}. Conversely, another type of work employs implicit conversion \cite{li2022bevformer,jiang2023polarformer}, which implicitly models depth through a cross-attention mechanism and extracts BEV features from image features. Remarkably, the performance of camera-based BEV perception in downstream tasks is now on par with that of LiDAR-based methods \cite{li2023bevdepth}. In contrast, occupancy perception can be regarded as an extension of BEV perception. Occupancy perception constructs a 3D volumetric space instead of a 2D BEV plane, resulting in a more complete description of the 3D scene.

\subsubsection{3D Semantic Scene completion}
3D semantic scene completion (3D SSC) is the task of simultaneously estimating the geometry and semantics of a 3D environment within a given range from limited observations, which requires imagining the complete 3D content of occluded objects and scenes. From a task content perspective, 3D semantic scene completion \cite{monoscene,sscbench,mei2023camera,yao2023depthssc,miao2023occdepth} aligns with semantic occupancy perception \cite{ming2024occfusion,huang2023tri,ganesh2023soccdpt,zhang2023occformer,silva2024s2tpvformer}. 

Drawing on prior knowledge, humans excel at estimating the geometry and semantics of 3D environments and occluded regions, but this is more challenging for computers and machines \cite{roldao20223d}. SSCNet \cite{song2017semantic} first raised the problem of semantic scene completion and tried to address it via a convolutional neural network. Early 3D SSC research mainly dealt with static indoor scenes \cite{song2017semantic,firman2016structured,chang2017matterport3d}, such as NYU \cite{nyuv2} and SUNCG \cite{song2017semantic} datasets. After the release of the large-scale outdoor benchmark SemanticKITTI \cite{semantickitti}, numerous outdoor SSC methods emerged. Among them, MonoScene \cite{monoscene} introduced the first monocular method for outdoor 3D semantic scene completion. It employs 2D-to-3D back projection to lift the 2D image and utilizes consecutive 2D and 3D UNets for semantic scene completion. In recent years, an increasing number of approaches have incorporated multi-camera and temporal information \cite{silva2024s2tpvformer,min2024multi,lyu20233dopformer} to enhance model comprehension of scenes and reduce completion ambiguity.

\subsubsection{3D Reconstruction from images}
3D reconstruction is a traditional but important topic in the computer vision and robotics communities \cite{hane2016dense,chen2024neuralrecon,tian2022high,leite2024fusing}. The objective of 3D reconstruction from images is to construct 3D of an object or scene based on 2D images captured from one or more viewpoints. Early methods exploited shape-from-shading \cite{durou2008numerical} or structure-from-motion \cite{schonberger2016structure}. Afterwards, the neural radiation field (NeRF) \cite{mildenhall2021nerf} introduced a novel paradigm for 3D reconstruction, which learned the density and color fields of 3D scenes, producing results with unprecedented detail and fidelity. However, such performance necessitates substantial training time and resources for rendering \cite{garbin2021fastnerf,reiser2021kilonerf,takikawa2021neural}, especially for high-resolution output. Recently, 3D Gaussian splatting (3D GS) \cite{kerbl20233d} has addressed this issue by redefining a paradigm-shifting approach to scene representation and rendering. Specifically, it represents scene representation with millions of 3D Gaussian functions in an explicit way, achieving faster and more efficient rendering \cite{chen2024survey}. 3D reconstruction emphasizes the geometric quality and visual appearance of the scene. In comparison, voxel-wise occupancy perception has lower resolution and visual appearance requirements, focusing instead on the occupancy distribution and semantic understanding of the scene.

\section{Methodologies}
\label{sec:Methodologies}

\begin{table*}[!t]   
\setlength{\belowcaptionskip}{0pt}
	\scriptsize
	\setlength{\tabcolsep}{0.003\linewidth}

	\definecolor{color1}{rgb}{0.0, 0.5, 0.69}
	\definecolor{color2}{rgb}{1.0, 0.22, 0.0}
	\definecolor{color3}{rgb}{0.0, 0.8, 0.6}
	\definecolor{color4}{rgb}{0.93, 0.53, 0.18}
        \definecolor{color5}{rgb}{0.93, 0.13, 0.0}
	\newcommand{\semloss}[1]{\textcolor{color1}{#1}}
	\newcommand{\geoloss}[1]{\textcolor{color2}{#1}}
	\newcommand{\csyloss}[1]{\textcolor{color3}{#1}}
        \newcommand{\semgeoloss}[1]{\textcolor{color4}{#1}}
        \newcommand{\distillloss}[1]{\textcolor{color5}{#1}}
	\newcommand{\AdvLoss}[1]{}

        \newcommand{\venue}[1]{\textit{#1}}
        \newcommand{\cam}[0]{C}
        \newcommand{\lidar}[0]{L}
        \newcommand{\radar}[0]{R}
        \newcommand{\tex}[0]{T}

        \newcommand{\prediction}[0]{P}
        \newcommand{\forecasting}[0]{F}
        \newcommand{\openvocabprediction}[0]{OP}
        \newcommand{\PanopticOccupancy}[0]{PO}
        \newcommand{\occupancy}[0]{O}

	\def\mystrut{\rule{0pt}{1.5\normalbaselineskip}}
	
	\centering
         \caption{\textbf{3D occupancy perception methods for autonomous driving.}}
	\rowcolors[]{3}{black!5}{white}
	\begin{tabular}{c|c | c |c c    c c c  | c |  c  c | c  c  c | c c}
		\toprule
		
		Method& 
		Venue& 
		  Modality &  
		 \multicolumn{5}{c|}{Design Choices}& 
		Task & 
		\multicolumn{2}{c|}{Training} & 
		\multicolumn{3}{c|}{Evaluation Datases} & 
		\multicolumn{2}{c}{Open Source} \\
		
		
		&
		& 
		& 
		\rotatebox{90}{Feature Format}  & 
		\rotatebox{90}{Multi-Camera} & 
		\rotatebox{90}{Multi-Frame} & 
		\rotatebox{90}{Lightweight Design} & 
		\rotatebox{90}{Head} &  
		   &
		\rotatebox{90}{Supervision} &  
		\rotatebox{90}{Loss} & 
		\rotatebox{90}{SemanticKITTI \cite{semantickitti}} & 
		\rotatebox{90}{Occ3D-nuScenes \cite{occ3d}} & 
		\rotatebox{90}{Others} & 
		\rotatebox{90}{Code} & 
		\rotatebox{90}{Weight} 
            \\
		\midrule

            LMSCNet \cite{roldao2020lmscnet}  & \venue{3DV 2020} & \lidar	& BEV	&-	&-	&2D Conv	&3D Conv	& \prediction	&Strong &	\semloss{CE} & \checkmark	&-	&-	&\checkmark	&\checkmark  \\
            S3CNet \cite{cheng2021s3cnet}	&\venue{CoRL 2020}	&\lidar	&BEV$+$Vol	&-	&-	&Sparse Conv	&2D \&3D Conv	&\prediction	&Strong	&\semloss{CE}, \semloss{PA}, \geoloss{BCE}	&\checkmark	&-	&-	&-	&-  \\
            
            DIFs \cite{rist2021semantic} \mystrut &	\venue{T-PAMI 2021}	&\lidar	&BEV	&-	&-	&2D Conv	&MLP	&\prediction	&Strong	&\semloss{CE}, \geoloss{BCE}, \csyloss{SC}	&\checkmark	&-	&-	&-	&-  \\
            
            MonoScene \cite{monoscene} \mystrut	&\venue{CVPR 2022}	&\cam	&Vol	&-	&-	&-	&3D Conv	&\prediction	&Strong	&\semloss{CE}, \semloss{FP}, \semgeoloss{Aff}	&\checkmark	&-	&-	&\checkmark	&\checkmark \\
            
            TPVFormer \cite{huang2023tri} \mystrut	&	\venue{CVPR 2023}&	\cam&	TPV&	\checkmark&	-&	TPV Rp&	MLP&	\prediction &	Strong&	\semloss{CE}, \semloss{LS}	&\checkmark	&-	&-	&\checkmark	&\checkmark \\
            VoxFormer \cite{li2023voxformer}	&\venue{CVPR 2023}	&\cam	&Vol	&-	&\checkmark	&-	&MLP	& \prediction	&Strong	&\semloss{CE}, \geoloss{BCE}, \semgeoloss{Aff}	&\checkmark	&-	&-	&\checkmark	&\checkmark  \\
            OccFormer \cite{zhang2023occformer}	&\venue{ICCV 2023}	&\cam	&BEV$+$Vol	&-	&-	&-	&Mask Decoder	& \prediction	&Strong	&\geoloss{BCE}, \semgeoloss{MC}	&\checkmark	&-	&-	&\checkmark	&\checkmark  \\
            OccNet \cite{openocc}	&\venue{ICCV 2023}	&\cam	&BEV$+$Vol	&\checkmark	&\checkmark	&-	&MLP	& \prediction	&Strong	&\semloss{Foc}	&-	&-	&OpenOcc \cite{openocc}	&\checkmark	&-  \\
            SurroundOcc \cite{surroundocc}	&\venue{ICCV 2023}	&\cam	&Vol	&\checkmark	&-	&-	&3D Conv	& \prediction	&Strong	&\semloss{CE}, \semgeoloss{Aff}	&\checkmark	&-	&SurroundOcc \cite{surroundocc}	&\checkmark	&\checkmark  \\
            OpenOccupancy \cite{openoccupancy}	&\venue{ICCV 2023}	&\cam$+$\lidar	&Vol 	&\checkmark	&-	&-	&3D Conv	&\prediction	&Strong	&\semloss{CE}, \semloss{LS}, \semgeoloss{Aff}	&-	&-	&OpenOccupancy \cite{openoccupancy}	&\checkmark	&-   \\
            NDC-Scene \cite{yao2023ndc} 	&\venue{ICCV 2023}	&\cam	&Vol	&-	&-	&-	&3D Conv	&\prediction	&Strong	&\semloss{CE}, \semloss{FP}, \semgeoloss{Aff}	&\checkmark	&-	&-	&\checkmark	&\checkmark \\
            Occ3D \cite{occ3d} 	&\venue{NeurIPS 2023}	&\cam	&Vol	&\checkmark	&-	&-	&MLP	& \prediction	&-	&-	&-	&\checkmark	&Occ3D-Waymo \cite{occ3d}	&-	&-   \\
            POP-3D \cite{vobecky2024pop}	&\venue{NeurIPS 2023}	&\cam$+$\tex$+$\lidar	&Vol 	&\checkmark	&-	&-	&MLP	&\openvocabprediction	&Semi	&\semloss{CE}, \semloss{LS}, \csyloss{MA}	&-	&-	&POP-3D	\cite{vobecky2024pop}&\checkmark	&\checkmark  \\
            OCF	\cite{ocfbench} &\venue{arXiv 2023}	&\lidar	&BEV/Vol 	&-	&\checkmark	&-	&3D Conv	&\prediction \&\forecasting	&Strong	&\geoloss{BCE}, \geoloss{SI}	&-	&-	&OCFBen \cite{ocfbench}	&\checkmark	&-   \\
            PointOcc \cite{zuo2023pointocc}	&\venue{arXiv 2023}	&\lidar	&TPV	&-	&-	&TPV Rp	&MLP	& \prediction	&Strong	&\semgeoloss{Aff}	&-	&-	&OpenOccupancy \cite{openoccupancy}	&\checkmark	&\checkmark   \\
            FlashOcc \cite{yu2023flashocc}	&\venue{arXiv 2023}	&\cam	&BEV	&\checkmark	&\checkmark	&2D Conv	&2D Conv	& \prediction	&-	&-	&-	&\checkmark	&-	&\checkmark	&\checkmark   \\
            OccNeRF \cite{zhang2023occnerf}	&\venue{arXiv 2023}	&\cam	&Vol 	&\checkmark	&\checkmark	&-	&3D Conv	& \prediction	&Self	&\semloss{CE}, \csyloss{Pho} &-	&\checkmark	&-	&\checkmark	&\checkmark    \\

            Vampire \cite{xu2024regulating} \mystrut &\venue{AAAI 2024}	&\cam	&Vol 	&\checkmark	&-	&-	&MLP+T	& \prediction	&Weak	&\semloss{CE}, \semloss{LS}	&-	&\checkmark	&-	&\checkmark	&\checkmark      \\
            FastOcc \cite{hou2024fastocc}	&\venue{ICRA 2024}	&\cam	&BEV	&\checkmark	&-	&2D Conv	&MLP	& \prediction	&Strong	&\semloss{CE}, \semloss{LS}, \semloss{Foc}, \geoloss{BCE}, \semgeoloss{Aff}	&-	&\checkmark	&-	&-	&-       \\
            RenderOcc \cite{pan2023renderocc}	&\venue{ICRA 2024}	&\cam	&Vol 	&\checkmark	&\checkmark	&-	&MLP+T	&\prediction	&Weak	&\semloss{CE}, \geoloss{SIL}	&\checkmark	&\checkmark	&-	&\checkmark	&\checkmark       \\
            MonoOcc	\cite{zheng2024monoocc} &\venue{ICRA 2024}	&\cam	&Vol 	&-	&\checkmark	&-	&MLP	&\prediction	&Strong	&\semloss{CE}, \semgeoloss{Aff}	&\checkmark	&-	&-	&\checkmark	&\checkmark      \\
            COTR \cite{ma2023cotr}	&\venue{CVPR 2024}	&\cam	&Vol 	&\checkmark	&\checkmark	&-	&Mask Decoder	&\prediction	&Strong	&\semloss{CE}, \semgeoloss{MC}	&-	&\checkmark	&-	&-	&-   \\
            Cam4DOcc \cite{cam4docc}	&\venue{CVPR 2024}	&\cam	&Vol 	&\checkmark	&\checkmark	&-	&3D Conv	& \prediction \&\forecasting	&Strong	&\semloss{CE}	&-	&-	&Cam4DOcc \cite{cam4docc}	&\checkmark	&\checkmark    \\
            PanoOcc \cite{wang2023panoocc}	&\venue{CVPR 2024}	&\cam	&Vol 	&\checkmark	&\checkmark	&-	&MLP, DETR	&\PanopticOccupancy	&Strong	&\semloss{Foc}, \semloss{LS}	&-	&\checkmark	&nuScenes \cite{nuscenes}	&\checkmark	&\checkmark     \\
            SelfOcc \cite{huang2023selfocc}	&\venue{CVPR 2024}	& \cam	&BEV/TPV	&\checkmark	&-	&-	&MLP+T	&\prediction	&Self	&\csyloss{Pho}	&\checkmark	&\checkmark	&-	&\checkmark	&\checkmark      \\
            Symphonies \cite{jiang2023symphonize}	&\venue{CVPR 2024}	&\cam	&Vol  &	-	&-	&-	&3D Conv	&\prediction	&Strong	&\semloss{CE}, \semgeoloss{Aff}	&\checkmark	&-	&SSCBench \cite{sscbench}	&\checkmark	&\checkmark      \\
            HASSC \cite{wang2024not}	&\venue{CVPR 2024}	&\cam	&Vol	&\checkmark	&\checkmark	&-	&MLP	&\prediction	&Strong	&\semloss{CE}, \semgeoloss{Aff}, \distillloss{KD}	&\checkmark	&-	&-	&-	&-\\
            SparseOcc \cite{tang2024sparseocc} 	&\venue{CVPR 2024}	&\cam	&Vol	&\checkmark	&-	&Sparse Rp	&Mask Decoder	&\prediction	&Strong	&-	&\checkmark	&-	&OpenOccupancy \cite{openoccupancy}	&-	&- \\
            MVBTS \cite{han2024boosting}	&\venue{CVPR 2024}	&\cam	&Vol	&\checkmark	&\checkmark	&-	&MLP+T	&\prediction	&Self	&\csyloss{Pho}, \distillloss{KD}	&-	&-	&KITTI-360 \cite{kitti360}	&\checkmark	&- \\
            DriveWorld \cite{min2024driveworld}	&\venue{CVPR 2024}	&\cam	&BEV	&\checkmark	&\checkmark	&-	&3D Conv	&\prediction	&Strong	&\semloss{CE}	&-	&-	&OpenScene \cite{openscene}	&-	&- \\
            Bi-SSC \cite{xue2024bi} &\venue{CVPR 2024}	&\cam	&BEV	&\checkmark	&-	&-	&3D Conv	&\prediction	&Strong	&\semloss{CE}, \semgeoloss{Aff}	&\checkmark	&-	&-	&-	&- \\
            LowRankOcc \cite{zhao2024lowrankocc} &\venue{CVPR 2024}	&\cam	&Vol	&-	&-	&TRDR	&Mask Decoder	&\prediction	&Strong	&\semgeoloss{MC}	&\checkmark	&-	&SurroundOcc \cite{surroundocc}	&-	&- \\
             PaSCo \cite{cao2024pasco} &\venue{CVPR 2024}	&\lidar	&Vol	&-	&-	&-	&Mask Decoder	&\PanopticOccupancy	&Strong	&\semloss{CE}, \semloss{LS}, \semgeoloss{MC}	&\checkmark	&-	&SSCBench \cite{sscbench}, Robo3D \cite{kong2023robo3d}	&\checkmark	&\checkmark \\
            HTCL \cite{li2024hierarchical} &\venue{ECCV 2024}	&\cam	&Vol	&\checkmark	&\checkmark	&-	&3D Conv	&\prediction 	&Strong	&\semloss{CE}, \semgeoloss{Aff}	&\checkmark	&-	&OpenOccupancy \cite{openoccupancy} 	&\checkmark	&\checkmark \\
            OSP \cite{shi2024occupancy} &\venue{ECCV 2024}	&\cam	&Point	&\checkmark	&-	&-	&MLP	&\prediction 	&Strong	&\semloss{CE}	&-	&\checkmark	&-	&\checkmark	&\checkmark \\
            OccGen \cite{wang2024occgen}&\venue{ECCV 2024}	&\cam$+$\lidar	&Vol	&\checkmark	&-	&-	&3D Conv	&\prediction	&Strong	&\semloss{CE}, \semloss{LS}, \semgeoloss{Aff}	&\checkmark	&-	&OpenOccupancy \cite{openoccupancy}	&-	&-       \\
            Scribble2Scene \cite{wang2024label} &\venue{IJCAI 2024}	&\cam	&Vol	&-	&\checkmark	&-	&MLP	&\prediction 	&Weak	&\semloss{CE}, \semgeoloss{Aff}, \distillloss{KD}	&\checkmark	&-	&SemanticPOSS \cite{pan2020semanticposs}	&-	&- \\
            BRGScene	&\venue{IJCAI 2024}	&\cam	&Vol	&\checkmark	 &-	&-	&3D Conv	&\prediction	&Strong	&\semloss{CE}, \geoloss{BCE}, \semgeoloss{Aff}	&\checkmark	&-	&-	&\checkmark	&\checkmark \\
            Co-Occ \cite{pan2024co}	&\venue{RA-L 2024}	&\cam$+$\lidar	&Vol 	&\checkmark	&-	&-	&3D Conv	&\prediction	&Strong	&\semloss{CE}, \semloss{LS}, \csyloss{Pho}	&\checkmark	&-	&SurroundOcc \cite{surroundocc}	&-	&-       \\
            OccFusion \cite{ming2024occfusion}	&\venue{arXiv 2024}	&\cam$+$\lidar/\radar	&BEV+Vol 	&\checkmark	&-	&-	&MLP	&\prediction	&Strong	&\semloss{Foc}, \semloss{LS}, \semgeoloss{Aff} &-	&-	&SurroundOcc \cite{surroundocc}	&-	&-       \\
            HyDRa	\cite{wolters2024unleashing}&\venue{arXiv 2024}	&\cam$+$\radar	&BEV+PV	&\checkmark	&-	&-	&3D Conv	&\prediction	&-	&-	&-	&\checkmark	&-	&-	&-       \\

		\bottomrule
	\end{tabular}\\
	{\scriptsize {\bf Modality}: C - Camera; L - LiDAR; R - Radar; T - Text.}\\
         {\scriptsize {\bf Feature Format}: Vol  - Volumetric Feature; BEV - Bird's-Eye View Feature; PV - Perspective View Feature; TPV - Tri-Perspective View Feature; Point - Point Feature}.\\
         {\scriptsize {\bf Lightweight Design}: TPV Rp  - Tri-Perspective View Representation; Sparse Rp - Sparse Representation; TRDR - Tensor Residual Decomposition and Recovery}.\\
         {\scriptsize {\bf Head}: MLP+T  - Multi-Layer Perceptron followed by Thresholding.}\\
	{\scriptsize {\bf Task}: P  - Prediction; F  - Forecasting; OP  - Open-Vocabulary Prediction; PO  - Panoptic Occupancy}.\\
	{\scriptsize {\bf Loss}: \geoloss{[Geometric]} BCE - Binary Cross Entropy, SIL - Scale-Invariant Logarithmic, SI - Soft-IoU; \semloss{[Semantic]} CE - Cross Entropy, PA - Position Awareness, FP - Frustum Proportion, LS - Lovasz-Softmax, Foc - Focal; \semgeoloss{[Semantic and Geometric]} Aff - Scene-Class Affinity, MC - Mask Classification; \csyloss{[Consistency]} SC - Spatial Consistency, MA - Modality Alignment, Pho - Photometric Consistency; \distillloss{[Distillation]} KD - Knowledge Distillation.
	}
	 
	\label{tab:methods}
\end{table*}

Recent methods of occupancy perception for autonomous driving and their characteristics are detailed in Tab. \ref{tab:methods}. This table elaborates on the publication venue, input modality, network design, target task, network training and evaluation, and open-source status of each method. In this section, according to the modality of input data, we categorize occupancy perception methods into three types: LiDAR-centric occupancy perception, vision-centric occupancy perception, and multi-modal occupancy perception. Additionally, network training strategies and corresponding loss functions will be discussed.

\subsection{LiDAR-Centric Occupancy Perception}
\subsubsection{General Pipeline}
LiDAR-centric semantic segmentation \cite{li2023less,kong2023lasermix,tang2023prototransfer} only predicts the semantic categories for sparse points. In contrast, LiDAR-centric occupancy perception provides a dense 3D understanding of the environment, crucial to autonomous driving systems. For LiDAR sensing, the acquired point clouds have an inherently sparse nature and suffer from occlusion. This requires that LiDAR-centric occupancy perception not only address the sparse-to-dense occupancy reasoning of the scene, but also achieve the partial-to-complete estimation of objects \cite{ming2024occfusion}.

\begin{figure}[t]
\setlength{\belowcaptionskip}{0pt}
    \centering
    \includegraphics[width=1\linewidth]{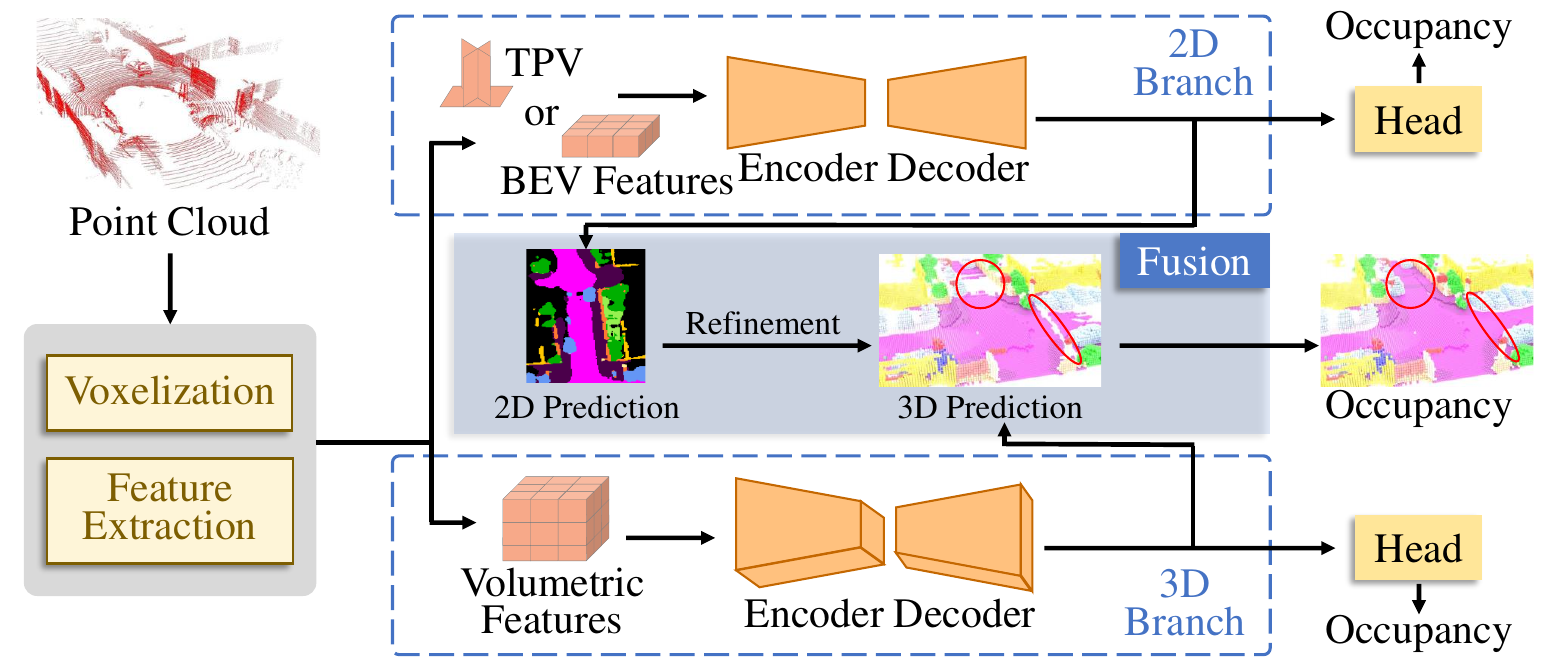}
    \caption{\textbf{Architecture for LiDAR-centric occupancy perception:} Solely the 2D branch \cite{rist2021semantic,zuo2023pointocc}, solely the 3D branch \cite{openoccupancy,roldao2020lmscnet,min2023occupancy}, and integrating both 2D and 3D branches \cite{cheng2021s3cnet}.}
    \label{fig:LiDAR-Centric Networks2}
\end{figure}

Fig. \ref{fig:LiDAR-Centric Networks2} illustrates the general pipeline of LiDAR-centric occupancy perception. The input point cloud first undergoes voxelization and feature extraction, followed by representation enhancement via an encoder-decoder module. Ultimately, the complete and dense occupancy of the scene is inferred. Specifically, given a point cloud $P\in \mathbb{R} ^{N\times 3}$, we generate a series of initial voxels and extract their features. These voxels are distributed in a 3D volume \cite{roldao2020lmscnet, cheng2021s3cnet,min2023occupancy,wang2024semantic}, a 2D BEV plane \cite{cheng2021s3cnet,rist2021semantic}, or three 2D tri-perspective view planes \cite{zuo2023pointocc}. 
This operation constructs the 3D feature volume or 2D feature map, denoted as $V_{\text{init}-3D} \in \mathbb{R} ^{X\times Y\times Z\times D}$ and $V_{\text{init}-2D}\in \mathbb{R} ^{X\times Y\times D}$ respectively. $N$ represents the number of points; $\left ( X, Y, Z \right ) $ are the length, width, and height; $D$ mean the feature dimensions of voxels. 
In addition to voxelizing in regular Euclidean space, PointOcc \cite{zuo2023pointocc} builds tri-perspective 2D feature maps in a cylindrical coordinate system. The cylindrical coordinate system aligns more closely with the spatial distribution of points in the LiDAR point cloud, where points closer to the LiDAR sensor are denser than those at farther distances. Therefore, it is reasonable to use smaller-sized cylindrical voxels for fine-grained modeling in nearby areas. The voxelization and feature extraction of point clouds can be formulated as:
\begin{equation}
    V_{\text{init}-2D\text{/}3D}=\Phi_{V}\left (\Psi _{V}\left ( P \right )   \right ), 
\end{equation}
where $\Psi _{V}$ stands for pillar or cubic voxelization. $\Phi_{V}$ is a feature extractor that extracts neural features of voxels (\textit{e.g.}, using PointPillars \cite{lang2019pointpillars}, VoxelNet \cite{zhou2018voxelnet}, and MLP) \cite{rist2021semantic,zuo2023pointocc}, or directly counts the geometric features of points within the voxel (\textit{e.g.}, mean, minimum, and maximum heights) \cite{cheng2021s3cnet,min2023occupancy}. Encoder and decoder can be various modules to enhance features. The final 3D occupancy inference involves applying convolution \cite{roldao2020lmscnet,cheng2021s3cnet,ocfbench} or MLP \cite{rist2021semantic,zuo2023pointocc,wang2024semantic} on the enhanced features to infer the occupied status $\left\{ 1,0\right\}$ of each voxel, and even estimate its semantic category:
\begin{equation}
    V=f_{\text{Conv/MLP}}\left ( ED\left ( V_{\text{init}-2D\text{/}3D} \right )  \right ) , 
\end{equation}
where $ED$ represents encoder and decoder.

\subsubsection{Information Fusion in LiDAR-Centric Occupancy}
Some works directly utilize a single 2D branch to reason about 3D occupancy, such as DIFs \cite{rist2021semantic} and PointOcc \cite{zuo2023pointocc}. In these approaches, only 2D feature maps instead of 3D feature volumes are required, resulting in reduced computational demand. However, a significant disadvantage is the partial loss of height information. In contrast, the 3D branch does not compress data in any dimension, thereby protecting the complete 3D scene. To enhance memory efficiency in the 3D branch, LMSCNet \cite{roldao2020lmscnet} turns the height dimension into the feature channel dimension. This adaptation facilitates the use of more efficient 2D convolutions compared to 3D convolutions in the 3D branch. Moreover, integrating information from both 2D and 3D branches can significantly refine occupancy predictions \cite{cheng2021s3cnet}.

S3CNet \cite{cheng2021s3cnet} proposes a unique late fusion strategy for integrating information from 2D and 3D branches. This fusion strategy involves a dynamic voxel fusion technique that leverages the results of the 2D branch to enhance the density of the output from the 3D branch. Ablation studies report that this straightforward and direct information fusion strategy can yield a $5$-$12\%$ performance boost in 3D occupancy perception.

\subsection{Vision-Centric Occupancy Perception}
\subsubsection{General Pipeline}
\begin{figure}[t]
\setlength{\belowcaptionskip}{0pt}
    \centering
    \includegraphics[width=1\linewidth]{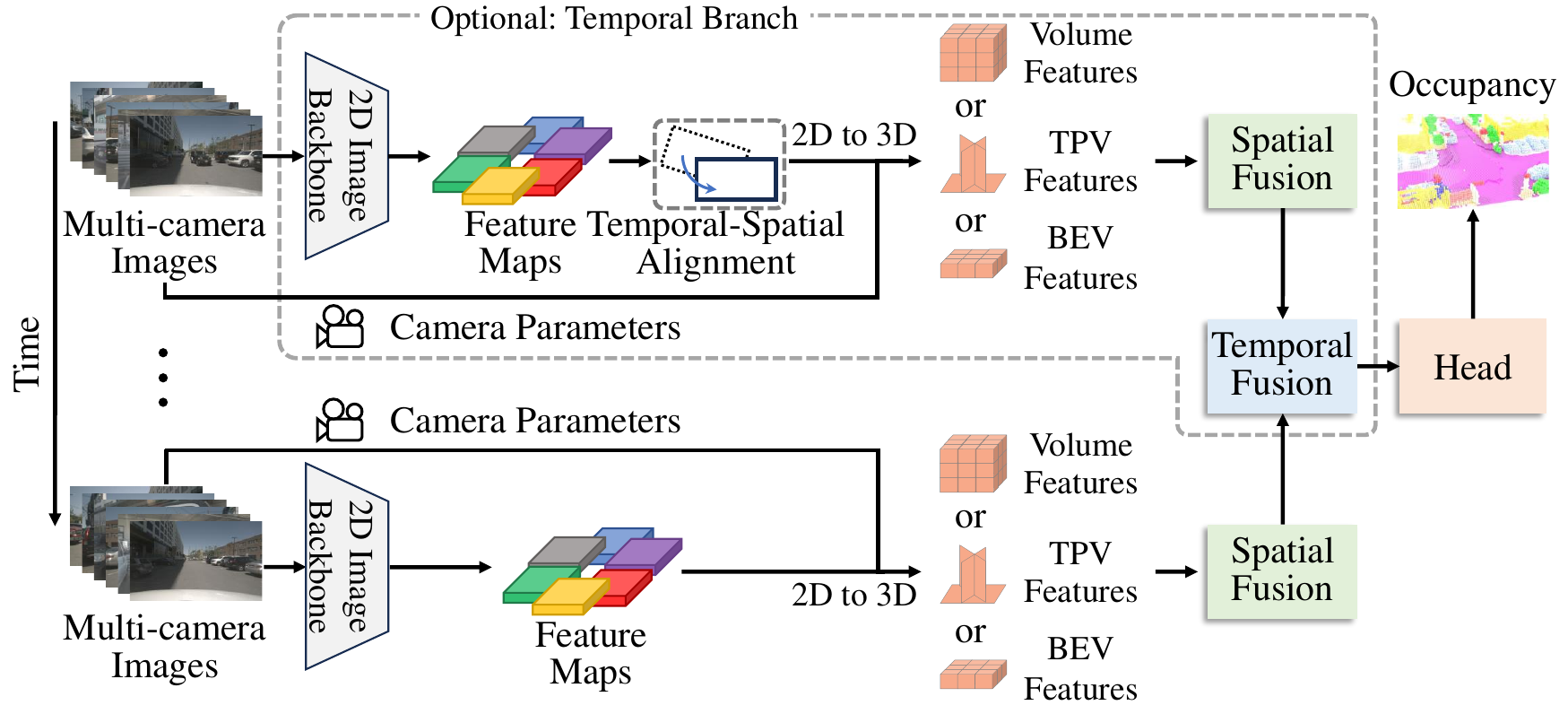}
    \caption{\textbf{Architecture for vision-centric occupancy perception:} Methods without temporal fusion \cite{gan2023simple,huang2023tri,occ3d,zhang2023occnerf,surroundocc,xu2024regulating,hou2024fastocc,pan2023renderocc,huang2023selfocc,boeder2024occflownet}; Methods with temporal fusion \cite{wang2023panoocc,openocc,cam4docc,silva2024s2tpvformer,yu2023flashocc,ma2023cotr}.}
    \label{fig:Vision-Centric Networks2}
\end{figure}

Inspired by Tesla's technology of the perception system for their autonomous vehicles \cite{WAD_2022}, vision-centric occupancy perception has garnered significant attention both in industry and academia. Compared to LiDAR-centric methods, vision-centric occupancy perception, which solely relies on camera sensors, represents a current trend. There are three main reasons: (i) Cameras are cost-effective for large-scale deployment on vehicles. (ii) RGB images capture rich environmental textures, aiding in the understanding of scenes and objects such as traffic signs and lane lines. (iii) The burgeoning advancement of deep learning technologies enables a possibility to achieve 3D occupancy perception from 2D vision. Vision-centric occupancy perception can be divided into monocular solutions \cite{monoscene,li2023voxformer,mei2023camera,yao2023depthssc,ganesh2023soccdpt,zhang2023occformer,zheng2024monoocc,jiang2023symphonize,tan2023ovo,shi2024panossc} and multi-camera solutions \cite{openocc,gan2023simple,huang2023tri,zhang2023occnerf,miao2023occdepth,min2024multi,yu2023flashocc,xu2024regulating,lu2023octreeocc,li2024bridging,tan2024geocc}. Multi-camera perception, which covers a broader field of view, follows a general pipeline as shown in Fig. \ref{fig:Vision-Centric Networks2}. It begins by extracting front-view feature maps from multi-camera images, followed by a 2D-to-3D transformation, spatial information fusion, and optional temporal information fusion, culminating with an occupancy head that infers the environmental 3D occupancy.

Specifically, the 2D feature map $F_{2D} (u, v)$ from the RGB image forms the basis of the vision-centric occupancy pipeline. Its extraction leverages the pre-trained image backbone network $\Phi _{F}$, such as convolution-based ResNet \cite{he2016deep} and Transformer-based ViT \cite{dosovitskiy2020image}, $ F_{2D}\left ( u,v \right ) =\Phi_{F}\left ( I\left ( u,v \right )  \right ) $. $I$ denotes the input image, $\left ( u,v \right )$ are pixel coordinates. Since the front view provides only a 2D perspective, a 2D-to-3D transformation is essential to deduce the depth dimension that the front view lacks, thereby enabling 3D scene perception. The 2D-to-3D transformation is detailed next.

\subsubsection{2D-to-3D Transformation}
\label{sec:2D-to-3D Transformation}
\begin{figure}[t]
\setlength{\belowcaptionskip}{0pt}
  \centering  
  \subfloat[2D-to-3D transformation. This serves as the fundamental unit for constructing 3D data from 2D observations, typically by projection \cite{monoscene,gan2023simple,zhang2023occnerf,miao2023occdepth,han2024boosting,ming2024inversematrixvt3d}, back projection \cite{zhang2023occformer,yu2023flashocc,xu2024regulating,hou2024fastocc,tang2024sparseocc}, and cross attention \cite{wang2023panoocc,occ3d,surroundocc,lu2023octreeocc,liu2024fully}. ]{	
    \includegraphics[width=1\linewidth]{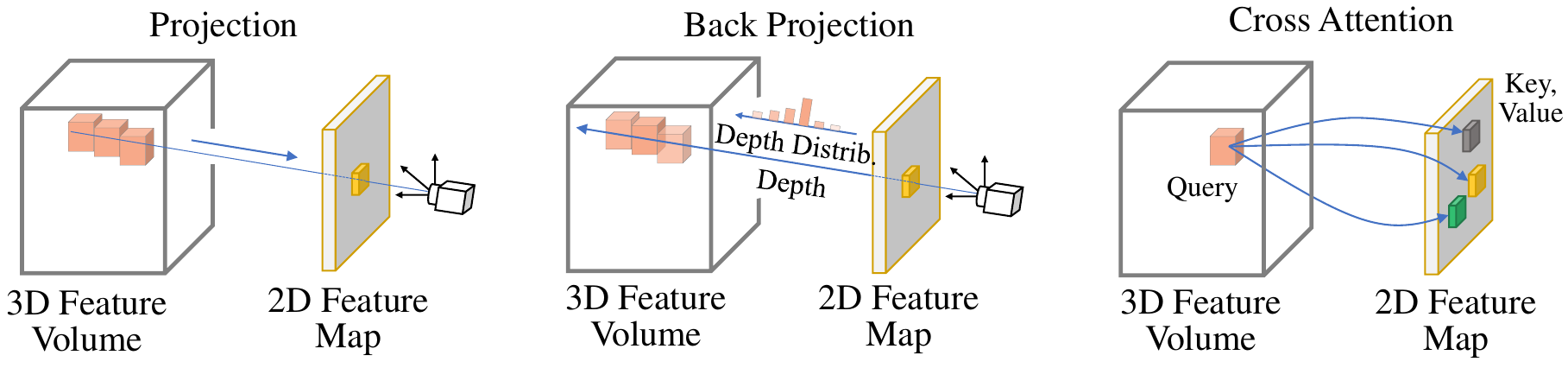}
	\label{fig:View Transformation}} \\
 \vspace{3pt}
    \subfloat[Spatial information fusion. In areas where multiple cameras have overlapping fields of view,  features from these cameras are fused through average \cite{zhang2023occnerf,miao2023occdepth,hou2024fastocc} or cross attention \cite{wang2023panoocc,huang2023tri,surroundocc,lu2023octreeocc,li2024viewformer}.]{	
    \includegraphics[width=0.9\linewidth]{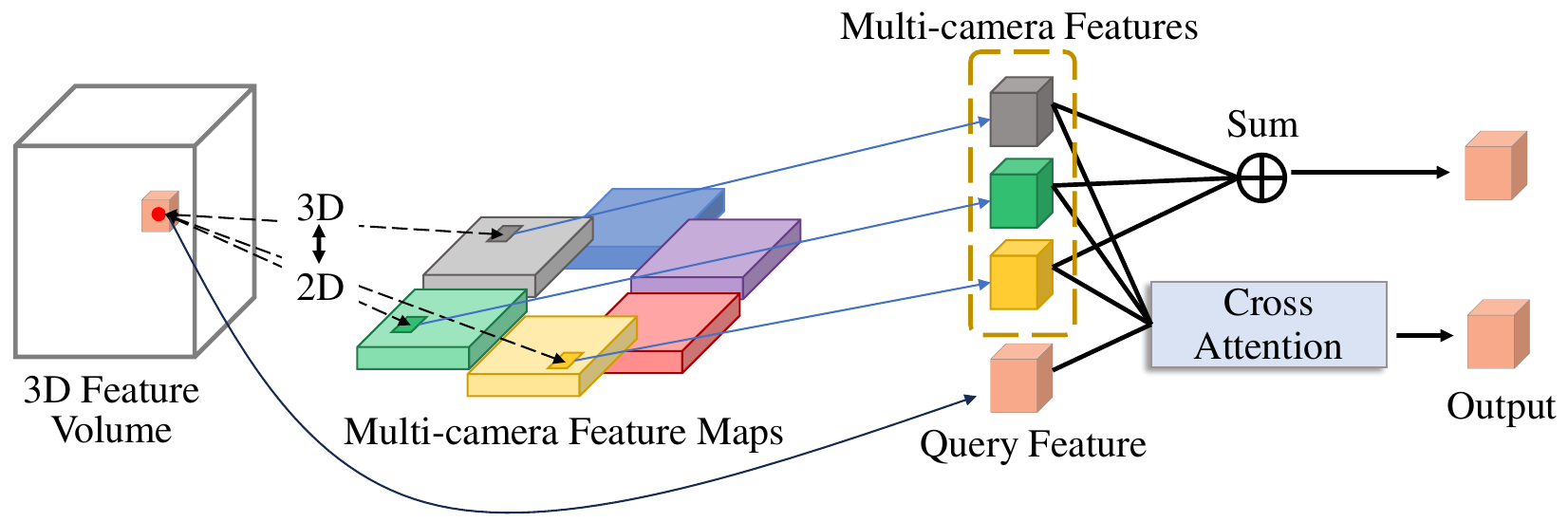}
	\label{fig:Multi-camera information integration}} \\
 \vspace{3pt}
    \subfloat[Temporal information fusion. Historical and current features undergo the temporal-spatial alignment, and then are fused via convolution \cite{wang2023panoocc} (see row $1$ in the feature fusion block), cross attention \cite{openocc,li2023voxformer,silva2024s2tpvformer,li2024viewformer} (row $2$), and adaptive mixing \cite{liu2024fully} (row $3$).]{	
    \includegraphics[width=1.0\linewidth]{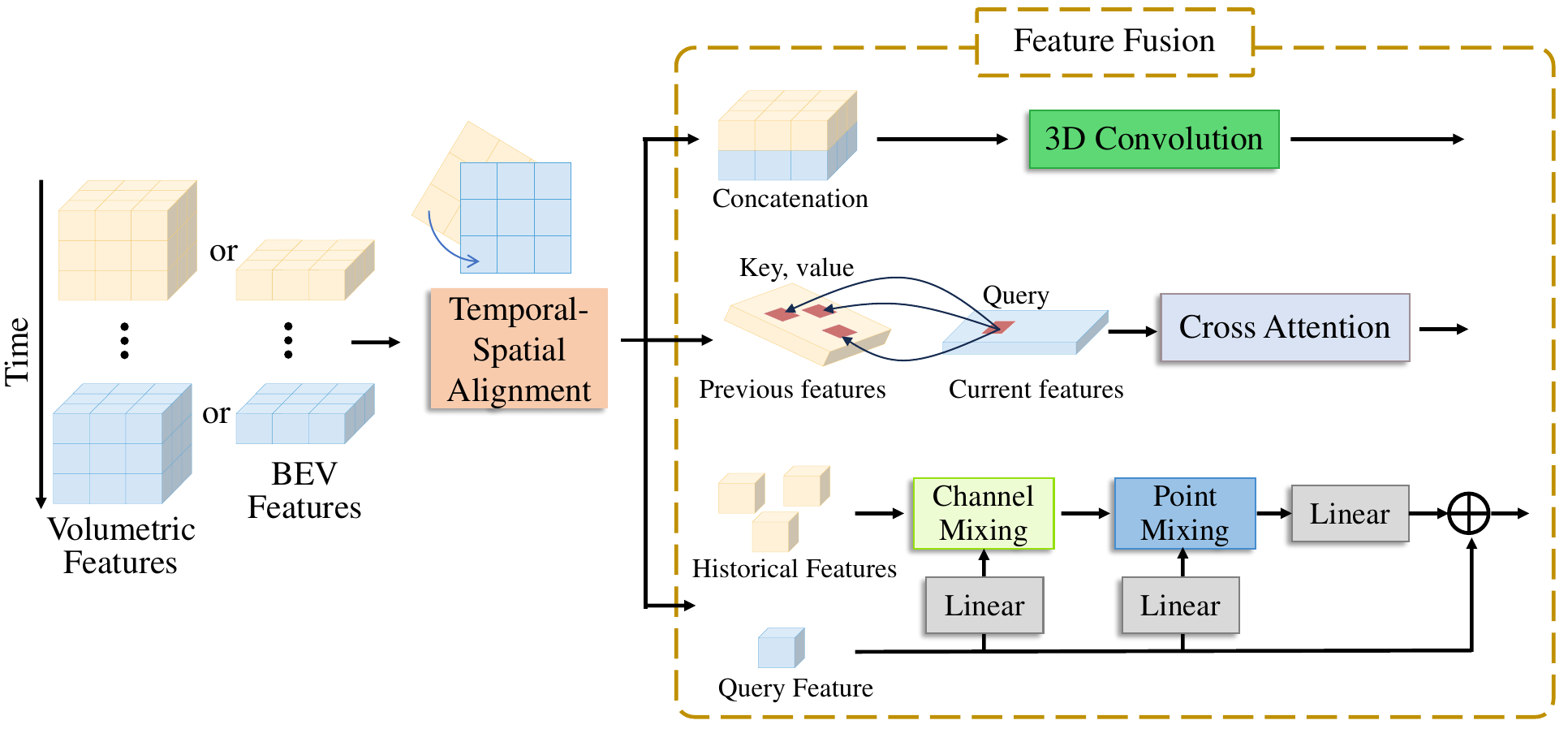}
	\label{fig:Historical Information Integration}}

  \caption{\textbf{Key components of vision-centric 3D occupancy perception.} Specifically, we present techniques for view transformation (\textit{i.e.}, 2D to 3D), multi-camera information integration (\textit{i.e.}, spatial fusion), and historical information integration (\textit{i.e.}, temporal fusion).}
  \label{fig:Key components of VOP}

\end{figure}

The transformation is designed to convert front-view features to BEV features \cite{min2024multi,yu2023flashocc}, TPV features \cite{huang2023tri}, or volumetric features \cite{li2023voxformer,surroundocc,ma2023cotr} to acquire the missing depth dimension of the front view. Notably, although BEV features are located on the top-view 2D plane, they can encode height information into the channel dimension of the features, thereby representing the 3D scene. The tri-perspective view projects the 3D space into three orthogonal 2D planes, so that each feature in the 3D space can be represented as a combination of three TPV features. The 2D-to-3D transformation is formulated as $F_{BEV/TPV/Vol}\left ( x^{\ast},y^{\ast},z^{\ast}  \right ) =\Phi _{T}\left ( F_{2D}\left ( u,v \right )  \right ) $, where $\left ( x,y,z \right )  $ represent the coordinates in the 3D space, $\ast$ means that the specific dimension may not exist in the BEV or TPV planes, $\Phi _{T}$ is the conversion from 2D to 3D. This transformation can be categorized into three types, characterized by using projection \cite{monoscene,gan2023simple,zhang2023occnerf,miao2023occdepth}, back projection \cite{zhang2023occformer,yu2023flashocc,xu2024regulating,hou2024fastocc}, and cross attention \cite{wang2023panoocc,occ3d,surroundocc,lu2023octreeocc,liu2024fully} technologies respectively. Taking the construction of volumetric features as an example, the process is illustrated in Fig. \ref{fig:View Transformation}.

(1) \textit{Projection}: It establishes a geometric mapping from the feature volume to the feature map. The mapping is achieved by projecting the voxel centroid in the 3D space onto the 2D front-view feature map through the perspective projection model $\Psi_{\rho }$ \cite{scaramuzza2011visual}, followed by performing sampling $\Psi _{S}$ by bilinear interpolation \cite{monoscene,gan2023simple,zhang2023occnerf,miao2023occdepth}. This projection process is formulated as:
\begin{equation}
    F_{Vol}\left ( x,y,z \right ) =\Psi _{S}\left ( F_{2D}\left (\Psi_{\rho } \left ( x,y,z,K,RT \right )  \right )  \right ) , 
\end{equation}
where $K$ and $RT$ are the intrinsics and extrinsics of the camera. However, the problem with the projection-based 2D-to-3D transformation is that along the line of sight, multiple voxels in the 3D space correspond to the same location in the front-view feature map. This leads to many-to-one mapping that introduces the ambiguity in the correspondence between 2D and 3D.

(2) \textit{Back Projection}: Back projection is the reverse process of projection. Similarly, it also utilizes perspective projection to establish correspondences between 2D and 3D. However, unlike projection, back projection uses the estimated depth $d$ of each pixel to calculate an accurate one-to-one mapping from 2D to 3D.
\begin{equation}
    F_{Vol}\left ( \Psi_{V}\left ( \Psi_{\rho }^{-1}\left ( u,v,d,K,RT \right )  \right )  \right )=F_{2D}\left ( u,v \right ), 
\end{equation}
where $\Psi_{\rho }^{-1}$ indicates the inverse projection function; $\Psi_{V}$ is voxelization. Since estimating the depth value may introduce errors, it is more effective to predict a discrete depth distribution $Dis$ along the optical ray rather than estimating a specific depth for each pixel \cite{zhang2023occformer,yu2023flashocc,xu2024regulating,hou2024fastocc}. That is, $F_{Vol}=F_{2D}\otimes Dis$, where $\otimes$ denotes outer product. This depth distribution-based re-projection, derived from LSS \cite{philion2020lift}, has significant advantages. On one hand, it can handle uncertainty and ambiguity in depth perception. For instance, if the depth of a certain pixel is unclear, the model can realize this uncertainty by the depth distribution. 
On the other hand, this probabilistic method of depth estimation provides greater robustness, particularly in a multi-camera setting. If corresponding pixels in multi-camera images have incorrect depth values to be mapped to the same voxel in the 3D space, their information might be unable to be integrated. In contrast, estimating depth distributions allows for information fusion with depth uncertainty, leading to more robustness and accuracy.

(3) \textit{Cross Attention}: The cross attention-based transformation aims to interact between the feature volume and the feature map in a learnable manner. Consistent with the attention mechanism \cite{vaswani2017attention}, each volumetric feature in the 3D feature volume acts as the query, and the key and value come from the 2D feature map. However, employing vanilla cross attention for the 2D-to-3D transformation requires considerable computational expense, as each query must attend to all features in the feature map. To optimize for GPU efficiency, many transformation methods \cite{wang2023panoocc,occ3d,surroundocc,lu2023octreeocc,liu2024fully} adopt deformable cross attention \cite{xia2022vision}, where the query interacts with selected reference features instead of all features in the feature map, therefore greatly reducing computation. Specifically, for each query, we project its 3D position $q$ onto the 2D feature map according to the given intrinsic and extrinsic. We sample some reference features around the projected 2D position $p$. These sampled features are then weighted and summed according to the deformable attention mechanism:
\begin{equation}
    F_{Vol}\left ( q \right )=\sum_{i=1}^{N_{head}}W_{i}\sum_{j=1}^{N_{key}}A_{ij}W_{ij}F_{2D}\left ( p+\bigtriangleup p_{ij} \right ) , 
\label{eq:deformable attention mechanism}
\end{equation}
where $\left ( W_{i}, W_{ij}\right ) $ are learnable weights, $A_{ij}$ denotes attention, $p+\bigtriangleup p_{ij}$ represents the position of the reference feature, and $\bigtriangleup p_{ij}$ indicates a learnable position shift.

Furthermore, there are some hybrid transformation methods that combine multiple 2D-to-3D transformation techniques. VoxFormer \cite{li2023voxformer} and SGN \cite{mei2023camera} initially compute a coarse 3D feature volume by per-pixel depth estimation and back projection, and subsequently refine the feature volume using cross attention. COTR \cite{ma2023cotr} has a similar hybrid transformation as VoxFormer and SGN, but it replaces per-pixel depth estimation with estimating depth distributions.

For TPV features, TPVFormer \cite{huang2023tri} achieves the 2D-to-3D transformation via cross attention. The conversion process differs slightly from that depicted in Fig. \ref{fig:View Transformation}, where the 3D feature volume is replaced by a 2D feature map in a specific perspective of three views. For BEV features, the conversion from the front view to the bird's-eye view can be achieved by cross attention \cite{min2024multi} or by back projection and then vertical pooling \cite{min2024multi,yu2023flashocc}.

\subsubsection{Information Fusion in Vision-Centric Occupancy}
In a multi-camera setting, each camera's front-view feature map describes a part of the scene. To comprehensively understand the scene, it is necessary to spatially fuse the information from multiple feature maps. Additionally, objects in the scene might be occluded or in motion. Temporally fusing feature maps of multiple frames can help reason about the occluded areas and recognize the motion status of objects.

(1) \textit{Spatial Information Fusion:} The fusion of observations from multiple cameras can create a 3D feature volume with an expanded field of view for scene perception. Within the overlapping area of multi-camera views, a 3D voxel in the feature volume will hit several 2D front-view feature maps after projection. There are two ways to fuse the hit 2D features: average \cite{zhang2023occnerf,miao2023occdepth,hou2024fastocc} and cross attention \cite{wang2023panoocc,huang2023tri,surroundocc,lu2023octreeocc}, as illustrated in Fig. \ref{fig:Multi-camera information integration}. The averaging operation calculates the mean of multiple features, which simplifies the fusion process and reduces computational costs. However, it assumes the equivalent contribution of different 2D perspectives to perceive the 3D scene. This may not always be the case, especially when certain views are occluded or blurry.

To address this problem, multi-camera cross attention is used to adaptively fuse information from multiple views. Specifically, its process can be regarded as an extension of Eq. \ref{eq:deformable attention mechanism} by incorporating more camera views. 
We redefine the deformable attention function as $DA\left ( q,p_{i},F_{2D\text{-}i } \right ) $, where $q$ is a query position in the 3D space, $p_{i}$ is its projection position on a specific 2D view, and $F_{2D\text{-}i }$ is the corresponding 2D front-view feature map. The multi-camera cross attention process can be formulated as:
\begin{equation}
    F_{Vol}\left ( q \right ) =\frac{1}{\left | \nu \right |  } \sum_{i\in \nu}DA\left ( q,p_{i},F_{2D\text{-}i } \right )  , 
\label{eq:multi-camera cross attention}
\end{equation}
where $F_{Vol}\left ( q \right ) $ represents the feature of the query position in the 3D feature volume, and $\nu$ denotes all hit views.

(2) \textit{Temporal Information Fusion:}
Recent advancements in vision-based BEV perception systems \cite{li2022bevformer,park2022time,liu2023petrv2} have demonstrated that integrating temporal information can significantly improve perception performance. Similarly, in vision-based occupancy perception, accuracy and reliability can be improved by combining relevant information from historical features and current perception inputs. The process of temporal information fusion consists of two components: temporal-spatial alignment and feature fusion, as illustrated in Fig. \ref{fig:Historical Information Integration}. The temporal-spatial alignment leverages pose information of the ego vehicle to spatially align historical features $F_{t-k}$ with current features. The alignment process is formulated as:
\begin{equation}
    F_{t-k}^{'}=\Psi _{S}\left ( T_{t-k\rightarrow t}\cdot F_{t-k} \right ), 
\end{equation}
where $T_{t-k\rightarrow t}$ is the transformation matrix that converts frame $t-k$ to the current frame $t$, involving translation and rotation; $\Psi_{S}$ represents feature sampling.

Once the alignment is completed, the historical and current features are fed to the feature fusion module to enhance the representation, especially to strengthen the reasoning ability for occlusion and the recognition ability of moving objects. There are three main streamlines to feature fusion, namely convolution, cross attention, and adaptive mixing. PanoOcc \cite{wang2023panoocc} concatenates the previous features with the current ones, then fuses them using a set of 3D residual convolution blocks. Many occupancy perception methods \cite{roldao20223d,li2023voxformer,silva2024s2tpvformer,zheng2024monoocc,li2024viewformer} utilize cross attention for fusion. The process is similar to multi-camera cross attention (refer to Eq. \ref{eq:multi-camera cross attention}), but the difference is that 3D-space voxels are projected to 2D multi-frame feature maps instead of multi-camera feature maps.
Moreover,  SparseOcc \cite{liu2024fully}\footnote{Concurrently, two works with the same name SparseOcc \cite{tang2024sparseocc,liu2024fully} explore sparsity in occupancy from different directions.} employs adaptive mixing \cite{liu2023sparsebev} for temporal information fusion. For the query feature of the current frame,  SparseOcc samples $S_{n}$ features from historical frames, and aggregates them through adaptive mixing. Specifically, the sampled features are multiplied by the channel mixing matrix $W_{C}$ and the point mixing matrix $W_{S_{n}}$, respectively. These mixing matrices are dynamically generated from the query feature $F_{q}$ of the current frame:
\begin{equation}
 W_{C/S_{n}}=\text{Linear} \left ( F_{q} \right ) \in \mathbb{R}^{C\times C} /\mathbb{R}^{S_{n}\times S_{n}}.
\end{equation}
The output of adaptive mixing is flattened, undergoes linear projection, and is then added to the query feature as residuals.

The features resulting from spatial and temporal information fusion are processed by various types of heads to determine 3D occupancy. These include convolutional heads, mask decoder heads, linear projection heads, and linear projection with threshold heads. Convolution-based heads \cite{min2024driveworld,cam4docc,monoscene,zhang2023occnerf,min2024multi,surroundocc,li2024bridging} consist of multiple 3D convolutional layers. Mask decoder-based heads \cite{zhang2023occformer,ma2023cotr,tang2024sparseocc,liu2024fully}, inspired by MaskFormer \cite{cheng2021maskformer} and Mask2Former \cite{cheng2021mask2former}, formalize 3D semantic occupancy prediction into the estimation of a set of binary 3D masks, each associated with a corresponding semantic category. Specifically, they compute per-voxel embeddings and assess per-query embeddings along with their related semantics. The final occupancy predictions are obtained by calculating the dot product of these two embeddings. Linear projection-based heads \cite{wang2023panoocc,huang2023tri,li2023voxformer,occ3d,mei2023camera,zheng2024monoocc,wang2024not} leverage lightweight MLPs on the dimension of feature channels to produce occupied status and semantics. Furthermore, for the occupancy methods\cite{xu2024regulating,pan2023renderocc,huang2023selfocc,han2024boosting,boeder2024occflownet} based on NeRF \cite{mildenhall2021nerf}, their occupancy heads use two separate MLPs ($\text {MLP}_{\sigma }$, $\text {MLP}_{s }$) to estimate the density volume $V_{\sigma}$ and the semantic volume $V_{S}$. Then the occupied voxels are selected based on a given confidence threshold $\tau $, and their semantic categories are determined based on $V_{S}$:
\begin{equation}
V\left ( x,y,z \right )=\begin{cases}
 \text{argmax}\left ( V_{S}\left ( x,y,z \right ) \right )  & \text{ if } V_{\sigma }\left ( x,y,z \right )\ge \tau  \\
 \text{empty} & \text{ if } V_{\sigma }\left ( x,y,z \right )<  \tau   ,
\end{cases} 
\end{equation}
where $\left ( x,y,z \right )$ represent 3D coordinates.

\begin{figure}[t]
\setlength{\belowcaptionskip}{0pt}
    \centering
    \includegraphics[width=1\linewidth]{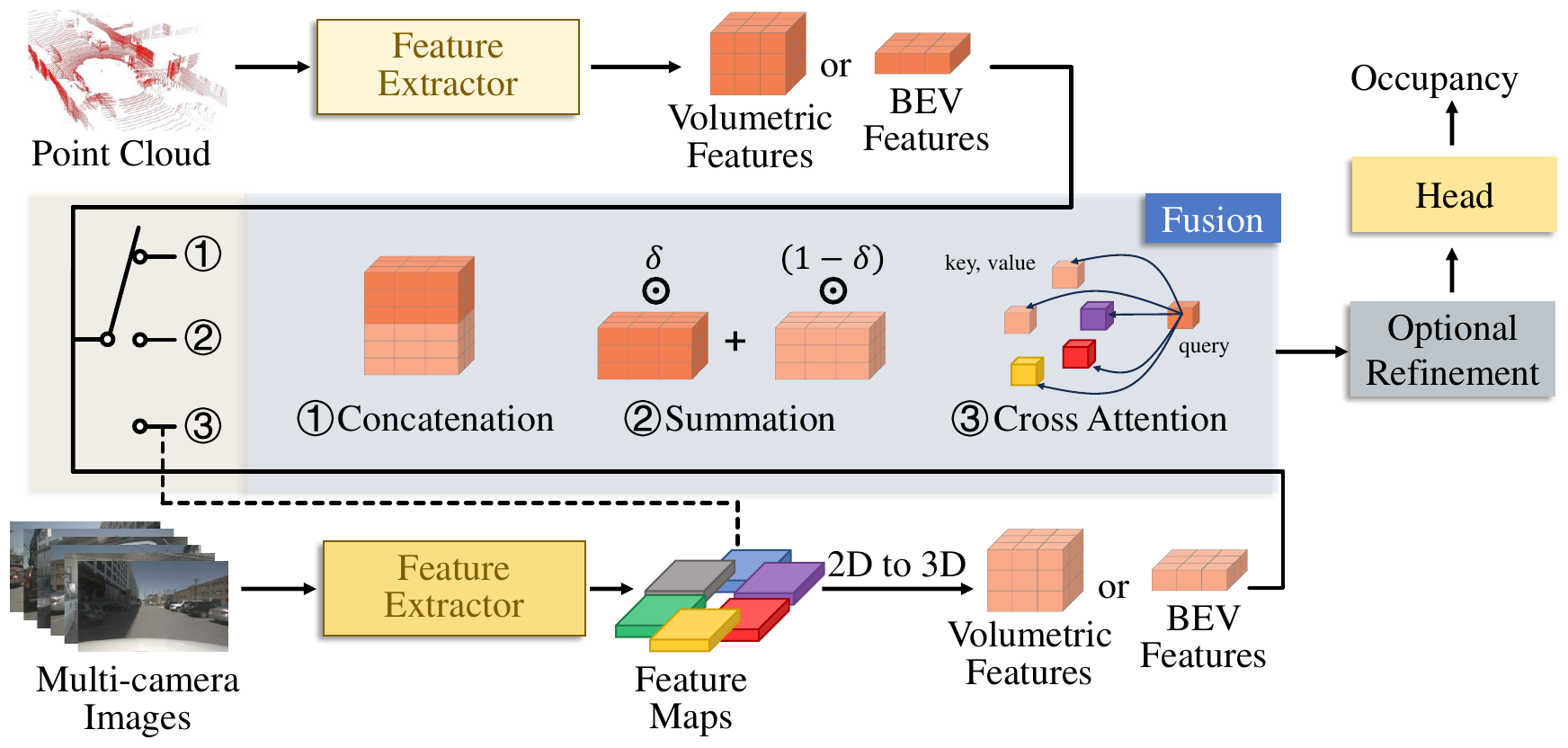}
    \caption{\textbf{Architecture for multi-modal occupancy perception:} Fusion of information from point clouds and images \cite{openoccupancy,ming2024occfusion,sze2024real,pan2024co,wang2024occgen}. Dashed lines signify additional fusion of perspective-view feature maps \cite{wolters2024unleashing}. $\odot$ represents element-wise product. $\delta$ is a learnable weight.}
    \label{fig:multi-modal Networks2}
\end{figure}

\subsection{Multi-Modal Occupancy Perception}
\subsubsection{General Pipeline}
RGB images captured by cameras provide rich and dense semantic information but are sensitive to weather condition changes and lack precise geometric details. In contrast, point clouds from LiDAR or radar are robust to weather changes and excel at capturing scene geometry with accurate depth measurements. However, they only produce sparse features. Multi-modal occupancy perception can combine the advantages from multiple modalities, and mitigate the limitations of single-modal perception. Fig. \ref{fig:multi-modal Networks2} illustrates the general pipeline of multi-modal occupancy perception. Most multi-modal methods \cite{openoccupancy,ming2024occfusion,sze2024real,pan2024co} map 2D image features into 3D space and then fuse them with point cloud features. Moreover, incorporating 2D perspective-view features in the fusion process can further refine the representation \cite{wolters2024unleashing}. The fused representation is processed by an optional refinement module and an occupancy head, such as 3D convolution or MLP, to generate the final 3D occupancy predictions. The optional refinement module \cite{wang2024occgen} could be a combination of cross attention, self attention, and diffusion denoising \cite{ho2020denoising}.

\subsubsection{Information Fusion in Multi-Modal Occupancy}
There are three primary multi-modal information fusion techniques to integrate different modality branches: concatenation, summation, and cross attention.

(1) \textit{Concatenation:}
Inspired by BEVFusion \cite{liu2023bevfusion,liang2022bevfusion}, OccFusion \cite{ming2024occfusion} combines 3D feature volumes from different modalities through concatenating them along the feature channel, and subsequently applies convolutional layers. Similarly, RT3DSO \cite{sze2024real} concatenates the intensity values of 3D points and their corresponding 2D image features (via projection), and then feeds the combined data to convolutional layers. However, some voxels in 3D space may only contain features from either the point cloud branch or the vision branch. To alleviate this problem, CO-Occ \cite{pan2024co} introduces the geometric- and semantic-aware fusion (GSFusion) module, which identifies voxels containing both point-cloud and visual information. This module utilizes a K-nearest
neighbors (KNN) search \cite{cover1967nearest} to select the $k$ nearest neighbors of a given position in voxel space within a specific radius. For the $i$-th non-empty feature $FL_{i}$ from the point-cloud branch, its nearest visual branch features are represented as $\left \{ FV_{i1},\cdots, FV_{ik} \right \} $, and a learnable weight $\omega_{i}$ is acquired by linear projection:
\begin{equation}
\omega_{i}=\text{Linear}\left ( \text{Concat}\left ( FV_{i1},\cdots, FV_{ik} \right )  \right ).
\end{equation}
The resulting LiDAR-vision features are expressed as $FLV=\text{Concat}\left ( FV,FL,FL\cdot \omega \right ) $, where $\omega$ denotes the geometric-semantic weight from $\omega_{i}$.

(2) \textit{Summation:}
CONet \cite{openoccupancy} and OccGen \cite{wang2024occgen} adopt an adaptive fusion module to dynamically integrate the occupancy representations from camera and LiDAR branches. It leverages 3D convolution to process multiple single-modal representations to determine their fusion weight, subsequently applying these weights to sum the LiDAR-branch representation and camera-branch features.

(3) \textit{Cross Attention:}
HyDRa \cite{wolters2024unleashing} proposes the integration of multi-modal information in perspective-view (PV) and BEV representation spaces. Specifically, the PV image features are improved by the BEV point-cloud features using cross attention. Afterwards, the enhanced PV image features are converted into BEV visual representation with estimated depth. These BEV visual features are further enhanced by concatenation with BEV point-cloud features, followed by a simple Squeeze-and-Excitation layer \cite{hu2018squeeze}. Finally, the enhanced PV image features and enhanced BEV visual features are fused through cross attention, resulting in the final occupancy representation.

\subsection{Network Training}
We classify network training techniques mentioned in the literature based on their supervised training types. The most prevalent type is strongly-supervised learning, while others employ weak, semi, or self supervision for training. This section details these network training techniques and their associated loss functions. The 'Training' column in Tab. \ref{tab:methods} offers a concise overview of network training across various occupancy perception methods.

\subsubsection{Training with Strong Supervision}
Strongly-supervised learning for occupancy perception involves using occupancy labels to train occupancy networks. Most occupancy perception methods adopt this training manner \cite{wang2023panoocc,cam4docc,monoscene,roldao2020lmscnet,huang2023tri,zhang2023occformer,surroundocc,hou2024fastocc,zheng2024monoocc,ma2023cotr,wang2024semantic,li2024bridging}. The corresponding loss functions can be categorized as: geometric losses, which optimize geometric accuracy; semantic losses, which enhance semantic prediction; combined semantic and geometric losses, which encourage both better semantic and geometric accuracy; consistency losses, encouraging overall consistency; and distillation losses, transferring knowledge from the teacher model to the student model. Next, we will provide detailed descriptions.

Among geometric losses, Binary Cross-Entropy (BCE) Loss is the most commonly used \cite{cheng2021s3cnet,li2023voxformer,zhang2023occformer,rist2021semantic,hou2024fastocc}, which distinguishes empty voxels and occupied voxels. The BCE loss is formulated as:
\begin{equation}
\mathcal{L}_{BCE} =-\frac{1}{N_{V}}\sum_{i=0}^{N_{V}}\hat{V}_{i}log\left ( V_{i} \right ) -\left ( 1- \hat{V}_{i}\right )  log \left ( 1-V_{i} \right ) ,
\end{equation}
where $N_{V}$ is the number of voxels in the occupancy volume $V$. Moreover, there are two other geometric losses: scale-invariant logarithmic loss \cite{eigen2014depth} and soft-IoU loss \cite{huang2019batching}. SimpleOccupancy \cite{gan2023simple} calculates the logarithmic difference between the predicted and ground-truth depths as the scale-invariant logarithmic loss. This loss relies on logarithmic rather than absolute differences, therefore offering certain scale invariance. OCF \cite{ocfbench} employs the soft-IoU loss to better optimize Intersection over Union (IoU) and prediction confidence.

Cross-entropy (CE) loss is the preferred loss to optimize occupancy semantics \cite{monoscene,huang2023tri,jiang2023symphonize,wang2024not,pan2024co,li2024bridging}. It treats classes as independent entities, and is formally expressed as:
\begin{equation}
\mathcal{L}_{CE} =-\frac{1}{N_{C}}\sum_{i=0}^{N_{V}}\sum_{c=0}^{N_{C}}\omega_{c} \hat{V}_{ic}log\left ( \frac{e^{V_{ic}}}{ {\textstyle \sum_{c'}^{N_{C}}}e^{V_{ic'}} }  \right ) ,
\end{equation}
where $\left ( V, \hat{V}\right ) $ are the ground-truth and predicted semantic occupancy with $N_{C}$ categories. $\omega_{c}$ is a weight for a specific class $c$ according to the inverse of the class frequency. Notably, CE loss and BCE loss are also widely used in semantic segmentation \cite{chen2023adversarial,chen2022semformer}. Besides these losses, some occupancy perception methods employ other semantic losses commonly utilized in semantic segmentation tasks \cite{wu2024joint,yan2024decoupling}, such as Lovasz-Softmax loss \cite{berman2018lovasz} and Focal loss \cite{lin2017focal}. Furthermore, there are two specialized semantic losses: frustum proportion loss \cite{monoscene}, which provides cues to alleviate occlusion ambiguities from the perspective of the visual frustum, and position awareness loss \cite{li2019depth}, which leverages local semantic entropy to encourage sharper semantic and geometric gradients.

The losses that can simultaneously optimize semantics and geometry for occupancy perception include scene-class affinity loss \cite{monoscene} and mask classification loss \cite{cheng2021maskformer,cheng2021mask2former}. The former optimizes the combination of precision, recall, and specificity from both geometric and semantic perspectives. The latter is typically associated with a mask decoder head \cite{zhang2023occformer,ma2023cotr}. Mask classification loss, originating from MaskFormer \cite{cheng2021maskformer} and Mask2Former \cite{cheng2021mask2former}, combines cross-entropy classification loss and a binary mask loss for each predicted mask segment.

The consistency loss and distillation loss correspond to spatial consistency loss \cite{rist2021semantic} and Kullback–Leibler (KL) divergence loss \cite{kullback1951information}, respectively. Spatial consistency loss minimizes the Jenssen-Shannon divergence of semantic inference between a given point and some support points in space, thereby enhancing the spatial consistency of semantics. KL divergence, also known as relative entropy, quantifies how one probability distribution deviates from a reference distribution. HASSC \cite{wang2024not} adopts KL divergence loss to encourage student models to learn more accurate occupancy from online soft labels provided by the teacher model.

\subsubsection{Training with Other Supervisions}
Training with strong supervision is straightforward and effective, but requires tedious annotation for voxel-wise labels. In contrast, training with other types of supervision, such as weak, semi, and self supervision, is label-efficient.

(1) \textit{Weak Supervision:} It indicates that occupancy labels are not used, and supervision is derived from alternative labels. For example, point clouds with semantic labels can guide occupancy prediction. Specifically, Vampire \cite{xu2024regulating} and RenderOcc \cite{pan2023renderocc} construct density and semantic volumes, which facilitate the inference of semantic occupancy of the scene and the computation of depth and semantic maps through volumetric rendering. These methods do not employ occupancy labels. Alternatively, they project LiDAR point clouds with semantic labels onto the camera plane to acquire ground-truth depth and semantics, which then supervise network training. Since both strongly-supervised and weakly-supervised learning predict geometric and semantic occupancy, the losses used in strongly-supervised learning, such as cross-entropy loss, Lovasz-Softmax loss, and scale-invariant logarithmic loss, are also applicable to weakly-supervised learning.

(2) \textit{Semi Supervision:} It utilizes occupancy labels but does not cover the complete scene, therefore providing only semi supervision for occupancy network training. POP-3D \cite{vobecky2024pop} initially generates occupancy labels by processing LiDAR point clouds, where a voxel is recorded as occupied if it contains at least one LiDAR point, and empty otherwise. Given the sparsity and occlusions inherent in LiDAR point clouds, the occupancy labels produced in this manner do not encompass the entire space, meaning that only portions of the scene have their occupancy labelled. POP-3D employs cross-entropy loss and Lovasz-Softmax loss to supervise network training. Moreover, to establish the cross-modal correspondence between text and 3D occupancy, POP-3D proposes to calculate the L2 mean square error between language-image features and 3D-language features as the modality alignment loss.

\begin{table*}[t]
\centering
\caption{\textbf{Overview of 3D occupancy datasets with multi-modal sensors.} Ann.: Annotation. Occ.: Occupancy. C: Camera. L: LiDAR. R: Radar. D: Depth map. Flow: 3D occupancy flow. Datasets highlighted in light gray are meta datasets.}
\resizebox{1.0\textwidth}{!}{
\begin{tabular}{c|c|c|c|c|c|c|c|c|c|c}
\toprule
                          &                        &                                & \multicolumn{5}{c|}{Sensor Data}                                & \multicolumn{3}{c}{Annotation}   \\
\cline{4-11}
\multirow{-2}{*}{Dataset} & \multirow{-2}{*}{Year} & \multirow{-2}{*}{Meta Dataset} & Modalities & Scenes & Frames/Clips with Ann.   & 3D Scans & Images & w/ 3D Occ.? & Classes & w/ Flow? \\
\midrule
\rowcolor[HTML]{EFEFEF} 
KITTI \cite{kitti}                     & \textit{CVPR 2012}     & -                              & C+L       & 22     & 15K Frames               & 15K      & 15k    & \xmk        & 21      & \cmk     \\
\rowcolor[HTML]{EFEFEF} 
SemanticKITTI \cite{semantickitti}     & \textit{ICCV 2019}     & KITTI \cite{kitti}                          & C+L       & 22     & 20K Frames               & 43K      & 15k    & \cmk        & 28      & \xmk     \\
\rowcolor[HTML]{EFEFEF} 
nuScenes \cite{nuscenes}               & \textit{CVPR 2019}     & -                              & C+L+R    & 1,000  & 40K Frames               & 390K     & 1.4M   & \xmk        & 32      & \xmk     \\
\rowcolor[HTML]{EFEFEF} 
Waymo \cite{waymo}                     & \textit{CVPR 2020}     & -                              & C+L       & 1,150  & 230K Frames              & 230K     & 12M    & \xmk        & 23      & \cmk     \\
\rowcolor[HTML]{EFEFEF} 
KITTI-360 \cite{kitti360}              & \textit{TPAMI 2022}    & -                              & C+L       & 11     & 80K Frames               & 320K     & 80K    & \xmk        & 19      & \xmk     \\
\midrule
MonoScene-SemanticKITT \cite{monoscene} & \textit{CVPR 2022}     & SemanticKITTI \cite{semantickitti}, KITTI \cite{kitti}           & C          & -      & 4.6K Clips           & -        & -      & \cmk        & 19      & \xmk     \\
MonoScene-NYUv2 \cite{monoscene}        & \textit{CVPR 2022}     & NYUv2 \cite{nyuv2}                          & C+D       & -      & 1.4K Clips           & -        & -      & \cmk        & 10      & \xmk     \\
\midrule
SSCBench-KITTI-360 \cite{sscbench}       & \textit{arXiv 2023}          & KITTI-360 \cite{kitti360}                      & C          & 9      & -                     & -        & -      & \cmk        & 19      & \xmk     \\
SSCBench-nuScenes \cite{sscbench}         & \textit{arXiv 2023}          & nuScenes \cite{nuscenes}                       & C          & 850    & -                     & -        & -      & \cmk        & 16      & \xmk     \\
SSCBench-Waymo \cite{sscbench}          & \textit{arXiv 2023}          & Waymo \cite{waymo}                          & C          & 1,000  & -                     & -        & -      & \cmk        & 14      & \xmk     \\
\midrule
OCFBench-Lyft \cite{ocfbench}             & \textit{arXiv 2023}          & Lyft-Level-5 \cite{lyft}                   & L          & 180    & 22K Frames               & -        & -      & \cmk        & -       & \xmk     \\
OCFBench-Argoverse \cite{ocfbench}       & \textit{arXiv 2023}          & Argoverse \cite{argoverse}                      & L          & 89     & 13K Frames               & -        & -      & \cmk        & 17      & \xmk     \\
OCFBench-ApolloScape \cite{ocfbench}     & \textit{arXiv 2023}          & ApolloScape \cite{apolloscape}                   & L          & 52     & 4K Frames                & -        & -      & \cmk        & 25      & \xmk     \\
OCFBench-nuScenes \cite{ocfbench}        & \textit{arXiv 2023}          & nuScenes \cite{nuscenes}                      & L          & -      & -                     & -        & -      & \cmk        & 16      & \xmk     \\
\midrule
SurroundOcc \cite{surroundocc}               & \textit{ICCV 2023}     & nuScenes \cite{nuscenes}                      & C          & 1,000  & -                     & -        & -      & \cmk        & 16      & \xmk     \\
OpenOccupancy \cite{openoccupancy}             & \textit{ICCV 2023}     & nuScenes \cite{nuscenes}                      & C+L       & -      & 34K Frames               & -        & -      & \cmk        & 16      & \xmk     \\
OpenOcc \cite{openocc}                   & \textit{ICCV 2023}     & nuScenes \cite{nuscenes}                      & C          & 850    & 40K Frames               & -        & -      & \cmk        & 16      & \xmk     \\
\midrule
Occ3D-nuScenes \cite{occ3d}            & \textit{NeurIPS 2024}  & nuScenes \cite{nuscenes}                      & C          & 900    & 1K Clips, 40K Frames     & -        & -      & \cmk        & 16      & \xmk     \\
Occ3D-Waymo \cite{occ3d}              & \textit{NeurIPS 2024}  & Waymo \cite{waymo}                         & C          & 1,000  & 1.1K Clips, 200K Frames & -        & -      & \cmk        & 14      & \xmk     \\
Cam4DOcc \cite{cam4docc}                  & \textit{CVPR 2024}     & nuScenes \cite{nuscenes} + Lyft-Level5 \cite{lyft}          & C+L       & 1,030  & 51K Frames                   & -        & -      & \cmk        & 2       & \cmk     \\
OpenScene \cite{openscene}                 & \textit{CVPR 2024 Challenge} & nuPlan \cite{nuplan}                         & C          & -      & 4M Frames                & 40M      & -      & \cmk        & -       & \cmk     \\
\bottomrule
\end{tabular}
}
\label{tab:datasets}
\end{table*} 

(3) \textit{Self Supervision:} It trains occupancy perception networks without any labels. To this end, volume rendering \cite{mildenhall2021nerf} provides a self-supervised signal to encourage consistency across different views from temporal and spatial perspectives, by minimizing photometric differences. MVBTS \cite{han2024boosting} computes the photometric difference between the rendered RGB image and the target RGB image. However,  several other methods calculate this difference between the warped image (from the source image) and the target image \cite{gan2023simple,zhang2023occnerf,huang2023selfocc}, where the depth needed for the warping process is acquired by volumetric rendering. OccNeRF \cite{zhang2023occnerf} believes that the reason for not comparing rendered images is that the large scale of outdoor scenes and few view supervision would make volume rendering networks difficult to converge. Mathematically, the photometric consistency loss \cite{garg2016unsupervised} combines a L1 loss and an optional structured similarity (SSIM) loss \cite{wang2004image} to calculate the reconstruction error between the warped image $\hat{I}$ and the target image $I$:
\begin{equation}
\mathcal{L}_{Pho} =\frac{\alpha }{2}\left ( 1-\mathrm{SSIM}\left ( I,\hat{I}  \right )   \right ) +\left ( 1-\alpha  \right ) \left \| I,\hat{I}  \right \| _{1},
\end{equation}
where $\alpha$ is a hyperparameter weight. Furthermore, OccNeRF leverages cross-Entropy loss for semantic optimization in a self-supervised manner. The semantic labels directly come from pre-trained semantic segmentation models, such as a pre-trained open-vocabulary model Grounded-SAM \cite{ren2024grounded,kirillov2023segment,liu2023grounding}.

\section{Evaluation}
\label{sec:Evaluation}
In this section, we will provide the performance evaluation of 3D occupancy perception. First, the datasets and metrics commonly used for evaluation are introduced. Subsequently, we offer detailed performance comparisons and discussions on state-of-the-art 3D occupancy perception methods using the most popular datasets.

\subsection{Datasets and Metrics}
\subsubsection{Datasets}

There are a variety of datasets to evaluate the performance of occupancy prediction approaches, \textit{e.g.}, the widely used KITTI \cite{kitti}, nuScenes \cite{nuscenes}, and Waymo \cite{waymo}. However, most of the datasets only contain 2D semantic segmentation annotations, which is not practical for the training or evaluation of 3D occupancy prediction approaches. To support the benchmarks for 3D occupancy perception, many new datasets such as Monoscene \cite{monoscene}, Occ3D \cite{occ3d}, and OpenScene \cite{openscene} are developed based on the previous datasets like nuScenes and Waymo. A detailed summary of datasets is provided in Tab. \ref{tab:datasets}.

\paragraph{Traditional Datasets} Before the development of 3D occupancy based algorithms, KITTI \cite{kitti}, SemanticKITTI \cite{semantickitti}, nuScenes \cite{nuscenes}, Waymo \cite{waymo}, and KITTI-360 \cite{kitti360} are widely used benchmarks for 2D semantic perception methods. KITTI contains $\sim$15K annotated frames from $\sim$15K 3D scans across 22 scenes with camera and LiDAR inputs. SemanticKITTI extends KITTI with more annotated frames ($\sim$20K) from more 3D scans ($\sim$43K). nuScenes collects more 3D scans ($\sim$390K) from 1,000 scenes, resulting in more annotated frames ($\sim$40K) and supports extra radar inputs. Waymo and KITTI-360 are two large datasets with $\sim$230K and $\sim$80K frames with annotations, respectively, while Waymo contains more scenes (1000 scenes) than KITTI-360 does (only 11 scenes). The above datasets are the widely adopted benchmarks for 2D perception algorithms before the popularity of 3D occupancy perception. These datasets also serve as the meta datasets of benchmarks for 3D occupancy based perception algorithms.

\paragraph{3D Occupancy Datasets} The occupancy network proposed by Tesla has led the trend of 3D occupancy based perception for autonomous driving. However, the lack of a publicly available large dataset containing 3D occupancy annotations brings difficulty to the development of 3D occupancy perception. To deal with this dilemma, many researchers develop 3D occupancy datasets based on meta datasets like nuScenes and Waymo. Monoscene \cite{monoscene} supporting 3D occupancy annotations is created from SemanticKITTI plus KITTI datasets, and NYUv2 \cite{nyuv2} datasets. SSCBench \cite{sscbench} is developed based on KITTI-360, nuScenes, and Waymo datasets with camera inputs. OCFBench \cite{ocfbench} built on Lyft-Level-5 \cite{lyft}, Argoverse \cite{argoverse}, ApolloScape \cite{apolloscape}, and nuScenes datasets only contain LiDAR inputs. SurroundOcc \cite{surroundocc}, OpenOccupancy \cite{openoccupancy}, and OpenOcc \cite{openocc} are developed on nuScenes dataset. Occ3D \cite{occ3d} contains more annotated frames with 3D occupancy labels ($\sim$40K based on nuScenes and $\sim$200K frames based on Waymo). Cam4DOcc \cite{cam4docc} and OpenScene \cite{openscene} are two new datasets that contain large-scale 3D occupancy and 3D occupancy flow annotations. Cam4DOcc is based on nuScenes plus Lyft-Level-5 datasets, while OpenScene with $\sim$4M frames with annotations is built on a very large dataset nuPlan \cite{nuplan}.

\subsubsection{Metrics}
(1) \textit{Voxel-level Metrics:} Occupancy prediction without semantic consideration is regarded as class-agnostic perception. It focuses solely on understanding spatial geometry, that is, determining whether each voxel in a 3D space is occupied or empty. The common evaluation metric is voxel-level Intersection-over-Union (IOU), expressed as:
\begin{equation}
    \text{IoU}=\frac{TP}{TP+FP+FN}, 
\end{equation}
where $TP$, $FP$, and $FN$ represent the number of true positives, false positives, and false negatives. A true positive means that an actual occupied voxel is correctly predicted.

Occupancy prediction that simultaneously infers the occupation status and semantic classification of voxels can be regarded as semantic-geometric perception. In this context, the mean Intersection-over-Union (mIoU) is commonly used as the evaluation metric. The mIoU metric calculates the IoU for each semantic class separately and then averages these IoUs across all classes, excluding the 'empty' class:
\begin{equation}
    \text{mIoU}=\frac{1}{N_{C}} \sum_{i=1}^{N_{C}} \frac{TP_{i}}{TP_{i}+FP_{i}+FN_{i}},
\label{eq:miou}
\end{equation}
where $TP_{i}$, $FP_{i}$, and $FN_{i}$ are the number of true positives, false positives, and false negatives for a specific semantic category $i$. $N_{C}$ denotes the total number of semantic categories.

(2) \textit{Ray-level Metric:} Although voxel-level IoU and mIoU metrics are widely recognized \cite{cam4docc,zhang2023occnerf,miao2023occdepth,surroundocc,yu2023flashocc,zheng2024monoocc,ma2023cotr,huang2023selfocc}, they still have limitations. Due to unbalanced distribution and occlusion of LiDAR sensing, ground-truth voxel labels from accumulated LiDAR point clouds are imperfect, where the areas not scanned by LiDAR are annotated as empty. Moreover, for thin objects, voxel-level metrics are too strict, as a one-voxel deviation would reduce the IoU values of thin objects to zero. To solve these issues,  SparseOcc \cite{liu2024fully} imitates LiDAR's ray casting and proposes ray-level mIoU, which evaluates rays to their closest contact surface. This novel mIoU, combined with the mean absolute velocity error (mAVE), is adopted by the occupancy score (OccScore) metric \cite{openscene}. OccScore overcomes the shortcomings of voxel-level metrics while also evaluating the performance in perceiving object motion in the scene (\textit{i.e.}, occupancy flow).

The formulation of ray-level mIoU is consistent with Eq. \ref{eq:miou} in form but differs in application. The ray-level mIoU evaluates each query ray rather than each voxel. A query ray is considered a true positive if (i) its predicted class label matches the ground-truth class and (ii) the L1 error between the predicted and ground-truth depths is below a given threshold. The mAVE measures the average velocity error for true positive rays among $8$ semantic categories. The final OccScore is calculated as:
\begin{equation}
     \text{OccScore}=\text{mIoU}\times 0.9+\text{max}\left ( 1- \text{mAVE},0.0 \right ) \times 0.1.
\end{equation}

\subsection{Performance}
In this subsection, we will compare and analyze the performance accuracy and inference speed of various 3D occupancy perception methods. For performance accuracy, we discuss three aspects: overall comparison, modality comparison, and supervision comparison. The evaluation datasets used include SemanticKITTI, Occ3D-nuScenes, and SSCBench-KITTI-360.

\subsubsection{Perception Accuracy}

\begin{table*}[t!]
		\footnotesize
		\setlength{\tabcolsep}{0.004\linewidth}
		\caption{\textbf{3D occupancy prediction comparison (\%) on the SemanticKITTI test set \cite{semantickitti}.} Mod.: Modality. C: Camera. L: LiDAR. The IoU evaluates the performance in geometric occupancy perception, and the mIoU evaluates semantic occupancy perception.
  }
		\newcommand{\classfreq}[1]{{~\tiny(\semkitfreq{#1}\%)}}  %
		\centering
              \setlength{\tabcolsep}{0.45mm}{
		\begin{tabular}{l|>{\columncolor{green!10}}c|>{\columncolor[HTML]{EFEFEF}}c>{\columncolor[HTML]{EFEFEF}}c| c c c c c c c c c c c c c c c c c c c}
			\toprule
			Method
                & \makecell[c]{Mod.}
                & IoU 
			 & mIoU 
			& \rotatebox{90}{\textcolor{road}{$\blacksquare$} road}
			\rotatebox{90}{\ \ \ \classfreq{road}} 
			& \rotatebox{90}{\textcolor{sidewalk}{$\blacksquare$} sidewalk}
			\rotatebox{90}{\ \ \ \classfreq{sidewalk}}
			& \rotatebox{90}{\textcolor{parking}{$\blacksquare$} parking}
			\rotatebox{90}{\ \ \ \classfreq{parking}} 
			& \rotatebox{90}{\textcolor{othergrnd}{$\blacksquare$} other-grnd}
			\rotatebox{90}{\ \ \ \classfreq{otherground}} 
			& \rotatebox{90}{\textcolor{building}{$\blacksquare$}  building}
			\rotatebox{90}{\ \ \ \classfreq{building}} 
			& \rotatebox{90}{\textcolor{car}{$\blacksquare$}  car}
			\rotatebox{90}{\ \ \ \classfreq{car}} 
			& \rotatebox{90}{\textcolor{truck}{$\blacksquare$}  truck}
			\rotatebox{90}{\ \ \ \classfreq{truck}} 
			& \rotatebox{90}{\textcolor{bicycle}{$\blacksquare$}  bicycle}
			\rotatebox{90}{\ \ \ \classfreq{bicycle}} 
			& \rotatebox{90}{\textcolor{motorcycle}{$\blacksquare$} motorcycle}
			\rotatebox{90}{\ \ \ \classfreq{motorcycle}} 
			& \rotatebox{90}{\textcolor{otherveh}{$\blacksquare$}  other-veh.}
			\rotatebox{90}{\ \ \  \classfreq{othervehicle}} 
			& \rotatebox{90}{\textcolor{vegetation}{$\blacksquare$} vegetation}
			\rotatebox{90}{\ \ \ \classfreq{vegetation}} 
			& \rotatebox{90}{\textcolor{trunk}{$\blacksquare$}  trunk}
			\rotatebox{90}{\ \ \ \classfreq{trunk}} 
			& \rotatebox{90}{\textcolor{terrain}{$\blacksquare$} terrain}
			\rotatebox{90}{\ \ \ \classfreq{terrain}} 
			& \rotatebox{90}{\textcolor{person}{$\blacksquare$}  person}
			\rotatebox{90}{\ \ \ \classfreq{person}} 
			& \rotatebox{90}{\textcolor{bicyclist}{$\blacksquare$}  bicyclist}
			\rotatebox{90}{\ \ \ \classfreq{bicyclist}} 
			& \rotatebox{90}{\textcolor{motorcycl}{$\blacksquare$}  motorcyclist.}
			\rotatebox{90}{\ \ \ \classfreq{motorcyclist}} 
			& \rotatebox{90}{\textcolor{fence}{$\blacksquare$} fence}
			\rotatebox{90}{\ \ \ \classfreq{fence}} 
			& \rotatebox{90}{\textcolor{pole}{$\blacksquare$} pole}
			\rotatebox{90}{\ \ \ \classfreq{pole}} 
			& \rotatebox{90}{\textcolor{trafsign}{$\blacksquare$} traf.-sign}
			\rotatebox{90}{\ \ \ \classfreq{trafficsign}} 
			\\
			\midrule

            S3CNet \cite{cheng2021s3cnet} &L	&45.60 	&\bf 29.53	&42.00               	&22.50	&17.00	&7.90	&\bf52.20	&31.20	&6.70	&\bf41.50	&\bf45.00	&\bf16.10	&39.50	&\bf34.00	&21.20	&\bf45.90	&\bf35.80	&\bf16.00	&\bf31.30	&\bf31.00	&\bf24.30  \\
            LMSCNet \cite{roldao2020lmscnet}	&L &56.72	&17.62	&64.80 	&34.68 	&29.02	&4.62	&38.08	&30.89 	&1.47	&0.00	&0.00	&0.81	&41.31	&19.89 	&32.05	&0.00	&0.00	&0.00	&21.32	&15.01	&0.84 \\
            JS3C-Net \cite{yan2021sparse}	&L &56.60	&23.75	&64.70 	&39.90 	&34.90 	&\bf14.10 	&39.40 	&33.30 	&\bf7.20 	&14.40 	&8.80	&12.70 	&43.10 	&19.60 	&40.50	&8.00 	&5.10 	&0.40 	&30.40 	&18.90 	&15.90 \\
            DIFs \cite{rist2021semantic}	&L &\bf 58.90	&23.56	&\bf69.60 	&\bf44.50	&\bf41.80 	&12.70 	&41.30	&\bf35.40 	&4.70	&3.60 	&2.70	&4.70	&\bf43.80 	&27.40 	&\bf40.90	&2.40	&1.00	&0.00	&30.50	&22.10 	&18.50 \\
            \midrule
            OpenOccupancy \cite{openoccupancy}	&C$+$L &-&20.42 	&60.60 	&36.10	&29.00 	&\bf13.00 	&\bf38.40 	&33.80 	&4.70	&3.00 	&2.20	&5.90	&\bf41.50	&20.50 	&35.10	&0.80 	&2.30 	&\bf0.60 	&26.00 	&18.70 	&15.70  \\
            Co-Occ \cite{pan2024co} &C$+$L &-	&\bf 24.44 	&\bf72.00 	&\bf43.50 	&\bf42.50	&10.20	&35.10 	&\bf40.00 	&\bf6.40 	&\bf4.40 	&\bf3.30	&\bf8.80 	&41.20 	&\bf30.80 	&\bf40.80 	&\bf1.60 	&\bf3.30 	&0.40 	&\bf32.70	&\bf26.60	&\bf20.70 \\
            \midrule
            MonoScene \cite{monoscene}	&C &34.16	&11.08	&54.70 	&27.10 	&24.80 	&5.70 	&14.40 	&18.80 	&3.30 	&0.50 	&0.70 	&4.40 	&14.90 	&2.40 	&19.50 	&1.00 	&1.40 	&0.40 	&11.10 	&3.30 	&2.10  \\
            TPVFormer \cite{huang2023tri}	&C &34.25 	&11.26	&55.10 	&27.20 	&27.40 	&6.50 	&14.80 	&19.20 	&3.70 	&1.00 	&0.50 	&2.30 	&13.90	&2.60 	&20.40 	&1.10 	&2.40 	&0.30 	&11.00 	&2.90 	&1.50 \\
            OccFormer \cite{zhang2023occformer}	&C &34.53 	&12.32	&55.90 	&30.30 	&31.50 	&6.50 	&15.70 	&21.60 	&1.20 	&1.50 	&1.70 	&3.20 	&16.80 	&3.90 	&21.30 	&2.20 	&1.10 	&0.20 	&11.90 	&3.80 	&3.70  \\
            SurroundOcc \cite{surroundocc}	&C &34.72 	&11.86	&56.90 	&28.30 	&30.20 	&6.80 	&15.20 	&20.60 	&1.40 	&1.60 	&1.20 	&4.40 	&14.90 	&3.40 	&19.30 	&1.40 	&2.00 	&0.10 	&11.30 	&3.90 	&2.40  \\
            NDC-Scene \cite{yao2023ndc}	&C &36.19	&12.58	&58.12 	&28.05 	&25.31 	&6.53 	&14.90 	&19.13 	&4.77 	&1.93 	&2.07 	&6.69 	&17.94 	&3.49 	&25.01 	&\bf3.44 	&2.77 	&1.64 	&12.85 	&4.43 	&2.96   \\
            RenderOcc \cite{pan2023renderocc}	&C & -	&8.24	&43.64	&19.10	&12.54	&0.00	&11.59	&14.83	&2.47	&0.42	&0.17	&1.78	&17.61	&1.48	&20.01	&0.94	&3.20	&0.00	&4.71	&1.17	&0.88  \\
            Symphonies \cite{jiang2023symphonize}	&C &42.19 	&15.04	&58.40 	&29.30	&26.90 	&11.70 	&24.70 	&23.60 	&3.20	&3.60 	&\bf2.60 	&5.60 	&24.20 	&10.00 	&23.10 	&3.20 	&1.90 	&\bf2.00 	&16.10 	&7.70 	&8.00   \\
            Scribble2Scene \cite{wang2024label}	&C &42.60 	&13.33	&50.30 	&27.30	&20.60 	&11.30 	&23.70 	&20.10 	&5.60	&2.70 	&1.60 	&4.50 	&23.50 	&9.60 	&23.80 	&1.60 	&1.80 	&0.00 	&13.30 	&5.60 	&6.50   \\
            HASSC \cite{wang2024not} &C &42.87	&14.38	&55.30	&29.60	&25.90	&11.30	&23.10	&23.00	&9.80	&1.90	&1.50	&4.90	&24.80	&9.80	&26.50	&1.40	&3.00	&0.00	&14.30	&7.00 	&7.10  \\
            BRGScene \cite{li2024bridging}	&C &43.34	&15.36	&61.90	&31.20	&30.70	&10.70	&24.20	&22.80	&8.40	&3.40	&2.40	&6.10	&23.80	&8.40	&27.00	&2.90	&2.20	&0.50	&16.50	&7.00	&7.20  \\
            VoxFormer \cite{li2023voxformer}	&C & 44.15	&13.35	&53.57	&26.52	&19.69	&0.42	&19.54	&26.54	&7.26	&1.28	&0.56	&\bf7.81	&\bf26.10	&6.10	&\bf33.06	&1.93	&1.97	&0.00	&7.31	&9.15	&4.94   \\
            MonoOcc \cite{zheng2024monoocc}	&C &- 	&15.63	&59.10 	&30.90 	&27.10 	&9.80 	&22.90 	&23.90	&7.20	&\bf4.50 	&2.40	&7.70	&25.00	&9.80	&26.10	&2.80 	&\bf4.70	&0.60 	&16.90 	&7.30 	&\bf8.40   \\
            HTCL \cite{li2024hierarchical}	&C &44.23 	&17.09	&64.40 	&\bf 34.80 	&\bf 33.80 	&\bf 12.40 	&25.90 	& \bf 27.30	&\bf 10.80	&1.80 	&2.20	&5.40	&25.30	&\bf 10.80	&31.20	&1.10 	&3.10	&0.90 	&21.10 	&9.00 	&8.30   \\
            Bi-SSC \cite{xue2024bi}	&C &\bf 45.10 	&\bf 16.73	&\bf 63.40 	& 33.30 	& 31.70 	&11.20 	&\bf 26.60 	&25.00	&6.80	&1.80 	&1.00	&6.80	&\bf 26.10	&10.50	&28.90	&1.70 	&3.30	&1.00 	&\bf 19.40 	&\bf 9.30 	&\bf 8.40   \\
			\bottomrule
		\end{tabular}
  }
		\label{tab:semantikitt_performance}
	\end{table*}

\begin{table*}[t]
	\footnotesize
		\setlength{\tabcolsep}{0.004\linewidth}
		\caption{\textbf{3D semantic occupancy prediction comparison (\%) on the validation set of Occ3D-nuScenes \cite{occ3d}.} Sup. represents the supervised learning type. mIoU$^{\ast}$ is the mean Intersection-over-Union excluding the 'others' and 'other flat' classes. For fairness, all compared methods are vision-centric.
  }
	\begin{tabular}{l|>{\columncolor{green!10}}c|>{\columncolor[HTML]{EFEFEF}}c>{\columncolor[HTML]{EFEFEF}}c| c c c c c c c c c c c c c c c c c}
		\toprule
		Method
		 & Sup.  & mIoU & mIoU$^{\ast}$
        & \rotatebox{90}{\textcolor{nothers}{$\blacksquare$} others}
        
		& \rotatebox{90}{\textcolor{nbarrier}{$\blacksquare$} barrier}
		
		& \rotatebox{90}{\textcolor{nbicycle}{$\blacksquare$} bicycle}
		
		& \rotatebox{90}{\textcolor{nbus}{$\blacksquare$} bus}

		& \rotatebox{90}{\textcolor{ncar}{$\blacksquare$} car}

		& \rotatebox{90}{\textcolor{nconstruct}{$\blacksquare$} const. veh.}

		& \rotatebox{90}{\textcolor{nmotor}{$\blacksquare$} motorcycle}

		& \rotatebox{90}{\textcolor{npedestrian}{$\blacksquare$} pedestrian}

		& \rotatebox{90}{\textcolor{ntraffic}{$\blacksquare$} traffic cone}

		& \rotatebox{90}{\textcolor{ntrailer}{$\blacksquare$} trailer}

		& \rotatebox{90}{\textcolor{ntruck}{$\blacksquare$} truck}

		& \rotatebox{90}{\textcolor{ndriveable}{$\blacksquare$} drive. suf.}

		& \rotatebox{90}{\textcolor{nother}{$\blacksquare$} other flat}

		& \rotatebox{90}{\textcolor{nsidewalk}{$\blacksquare$} sidewalk}

		& \rotatebox{90}{\textcolor{nterrain}{$\blacksquare$} terrain}

		& \rotatebox{90}{\textcolor{nmanmade}{$\blacksquare$} manmade}

		& \rotatebox{90}{\textcolor{nvegetation}{$\blacksquare$} vegetation}

		\\
		\midrule

        SelfOcc (BEV) \cite{huang2023selfocc}	&Self	&6.76 	&7.66	&0.00 	&0.00 	&0.00 	&0.00 	&9.82	&0.00 	&0.00 	&0.00 	&0.00 	&0.00 	&6.97	&47.03 	&0.00 	&18.75 	&16.58 	&11.93 	&3.81 \\
        SelfOcc (TPV) \cite{huang2023selfocc}	&Self	& \bf7.97 	&9.03	&0.00 	&0.00 	&0.00	&0.00 	&10.03	&0.00	&0.00 	&0.00 	&0.00	&0.00	&\bf7.11 	&\bf52.96	&0.00	&\bf23.59 	&\bf25.16 	&11.97	&4.61 \\
        SimpleOcc \cite{gan2023simple} &Self 	&-	&7.99	&-	&0.67 	&\bf1.18	&3.21 	&7.63	&1.02	&\bf0.26 	&1.80 	&0.26 	&\bf1.07 	&2.81 	&40.44	&-	&18.30 	&17.01 	&13.42 	&10.84  \\
        OccNeRF \cite{zhang2023occnerf} 	&Self 	&-	&\bf 10.81	&-	&\bf0.83	&0.82	&\bf5.13 	&\bf12.49 	&\bf3.50 	&0.23	&\bf3.10 	&\bf1.84 	&0.52 	&3.90 	&52.62	&-	&20.81 	&24.75 	&\bf18.45 	&\bf13.19 \\
        \midrule
        RenderOcc \cite{pan2023renderocc}  	&Weak	&23.93	&-	&5.69 	&27.56 	&14.36 	&19.91 	&20.56 	&11.96 	&12.42 	&12.14 	&14.34 	&20.81 	&18.94 	&\bf68.85 	&33.35 	&\bf42.01 	&\bf43.94 	&17.36 	&\bf22.61 \\
        Vampire \cite{xu2024regulating}	&Weak	& \bf28.33	&-	&\bf7.48 	&\bf32.64 	&\bf16.15 	&\bf36.73 	&\bf41.44	&\bf16.59	&\bf20.64	&\bf16.55 	&\bf15.09 	&\bf21.02	&\bf28.47 	&67.96 	&\bf33.73 	&41.61 	&40.76 	&\bf24.53 	&20.26 \\
        \midrule
        OccFormer \cite{zhang2023occformer}	&Strong	&21.93	&-	&5.94 	&30.29 	&12.32 	&34.40 	&39.17	&14.44 	&16.45 	&17.22 	&9.27 	&13.90 	&26.36	&50.99 	&30.96 	&34.66 	&22.73	&6.76	&6.97 \\
        TPVFormer \cite{huang2023tri}	&Strong	&27.83	&-	&7.22 	&38.90 	&13.67	&40.78 	&45.90	&17.23	&19.99 	&18.85 	&14.30 	&26.69	&34.17	&55.65 	&35.47	&37.55	&30.70 	&19.40 	&16.78  \\
        Occ3D \cite{occ3d}	&Strong	&28.53	&-	&8.09	&39.33	&20.56	&38.29	&42.24 	&16.93	&24.52	&22.72 	&21.05	&22.98 	&31.11 	&53.33	&33.84	&37.98 	&33.23 	&20.79	&18.00 \\
        SurroundOcc \cite{surroundocc}	&Strong	&38.69	&-	&9.42 	&43.61 	&19.57 	&47.66 	&53.77 	&21.26 	&22.35 	&24.48 	&19.36 	&32.96 	&39.06 	&83.15 	&43.26 	&52.35 	&55.35 	&43.27 	&38.02  \\
        FastOcc \cite{hou2024fastocc}	&Strong		&40.75	&-	&12.86 	&46.58 	&29.93 	&46.07 	&54.09 	&23.74 	&31.10 	&30.68 	&28.52 	&33.08 	&39.69 	&83.33 	&44.65 	&53.90 	&55.46 	&42.61 	&36.50 \\
        FB-OCC \cite{li2023fb}	&Strong	&42.06	&-	&14.30 	&49.71 	&30.00 	&46.62 	&51.54 	&29.30	&29.13	&29.35	&30.48 	&34.97 	&39.36 	&83.07 	&47.16 	&55.62	&59.88 	&44.89 	&39.58 \\
        PanoOcc \cite{wang2023panoocc}	&Strong	&42.13 	&-	&11.67 	&50.48 	&29.64 	&49.44 	&55.52 	&23.29 	&33.26 	&30.55 	&30.99 	&34.43 &42.57 	&83.31 	&44.23	&54.40 	&56.04 	&45.94 	&40.40 \\
        COTR \cite{ma2023cotr}	&Strong	& \bf 46.21	&-	&\bf14.85 	&\bf53.25 	&\bf35.19 	&\bf50.83 	&\bf57.25 	&\bf35.36 	&\bf34.06 	&\bf33.54 	&\bf37.14 	&\bf38.99 	&\bf44.97 	&\bf84.46 	&\bf48.73 	&\bf57.60 	&\bf61.08 	&\bf51.61 	&\bf46.72 \\
		\bottomrule
	\end{tabular}
    \label{tab:occ3d_performance}

\end{table*}

\begin{table*}[t]
    \footnotesize
    \setlength{\tabcolsep}{0.0046\linewidth}
    \caption{\textbf{3D occupancy benchmarking results (\%) on the SSCBench-KITTI-360 test set.} The best results are in bold. OccFiner (Mono.) indicates that OccFiner refines the predicted occupancy from MonoScene.}
    \newcommand{\clsname}[2]{
            \rotatebox{90}{
            \hspace{-6pt}
            \textcolor{#2}{$\blacksquare$}
            \hspace{-6pt}
            \renewcommand\arraystretch{0.6}
            \begin{tabular}{l}
                #1                                       \\
                \hspace{-4pt} ~\tiny(\sscbkitfreq{#2}\%) \\
            \end{tabular}
    }}
    \renewcommand\arraystretch{1.2}
    \centering
        \begin{tabular}{l|>{\columncolor{gray!20}}c>{\columncolor{gray!20}}c|cccccccccccccccccc}
            \toprule
            \multicolumn{1}{c|}{Method}                                 &
            IoU
            &
            mIoU                                                        &
            \multicolumn{1}{c}{\clsname{car}{car}}                      &
            \multicolumn{1}{c}{\clsname{bicycle}{bicycle}}              &
            \multicolumn{1}{c}{\clsname{motorcycle}{motorcycle}}        &
            \multicolumn{1}{c}{\clsname{truck}{truck}}                  &
            \multicolumn{1}{c}{\clsname{other-veh.}{otherveh}}      &
            \multicolumn{1}{c}{\clsname{person}{person}}                &
            \multicolumn{1}{c}{\clsname{road}{road}}                    &
            \multicolumn{1}{c}{\clsname{parking}{parking}}              &
            \multicolumn{1}{c}{\clsname{sidewalk}{sidewalk}}            &
            \multicolumn{1}{c}{\clsname{other-grnd.}{othergrnd}}      &
            \multicolumn{1}{c}{\clsname{building}{building}}            &
            \multicolumn{1}{c}{\clsname{fence}{fence}}                  &
            \multicolumn{1}{c}{\clsname{vegetation}{vegetation}}        &
            \multicolumn{1}{c}{\clsname{terrain}{terrain}}              &
            \multicolumn{1}{c}{\clsname{pole}{pole}}                    &
            \multicolumn{1}{c}{\clsname{traf.-sign}{trafsign}}       &
            \multicolumn{1}{c}{\clsname{other-struct.}{otherstructure}} &
            \multicolumn{1}{c}{\clsname{other-obj.}{otherobject}}
            \\
            \midrule
            \multicolumn{21}{l}{\textit{LiDAR-Centric Methods}}                                     \\
            \hline
            SSCNet \cite{song2017semantic}                                        & \bf 53.58  & \bf 16.95        & \bf 31.95 & 0.00        & \bf 0.17        & \bf 10.29        & 0.00         & \bf 0.07        & \bf 65.70 & \bf 17.33 & \bf 41.24 & \bf 3.22        & \bf 44.41 & \bf 6.77        & \bf 43.72 & \bf 28.87 & \bf 0.78         & \bf 0.75        & \bf 8.69         & \bf 0.67         \\
            LMSCNet \cite{roldao2020lmscnet}                                      & 47.35           & 13.65        & 20.91        & 0.00        & 0.00        & 0.26         & 0.58         & 0.00        & 62.95        & 13.51        & 33.51        & 0.20        & 43.67        & 0.33        & 40.01        & 26.80        & 0.00         & 0.00        & 3.63         & 0.00         \\
            \specialrule{0.7pt}{0pt}{0pt}
            \multicolumn{21}{l}{\textit{Vision-Centric Methods}}                                     \\
            \hline
            GaussianFormer \cite{huang2024gaussianformer}                                  & 35.38 &12.92        &18.93 &1.02 &4.62 &18.07 &7.59 &3.35 &45.47 &10.89 &25.03 &5.32 &28.44 &5.68 &29.54 &8.62 &2.99 &2.32 &9.51 &5.14 \\
            MonoScene \cite{monoscene}                                  & 37.87        & 12.31        & 19.34        & 0.43        & 0.58        & 8.02         & 2.03         & 0.86        & 48.35        & 11.38        & 28.13        & 3.32        & 32.89        & 3.53        & 26.15        & 16.75        & 6.92         & 5.67        & 4.20         & 3.09         \\
            OccFiner (Mono.) \cite{shi2024occfiner} & 38.51  &13.29        & 20.78 &1.08 &1.03 &9.04 &3.58 &1.46 &53.47 &12.55 &31.27 &4.13 &33.75 &4.62 &26.83 &18.67 &5.04 &4.58 &4.05 &3.32         \\
            VoxFormer \cite{li2023voxformer}                                  & 38.76     & 11.91        & 17.84        & 1.16        & 0.89        & 4.56         & 2.06         & 1.63        & 47.01        & 9.67         & 27.21        & 2.89        & 31.18        & 4.97        & 28.99        & 14.69        & 6.51         & 6.92        & 3.79         & 2.43         \\
            TPVFormer \cite{huang2023tri}                                  & 40.22      & 13.64        & 21.56        & 1.09        & 1.37        & 8.06         & 2.57         & 2.38        & 52.99        & 11.99        & 31.07        & 3.78        & 34.83        & 4.80        & 30.08        & 17.52        & 7.46         & 5.86        & 5.48         & 2.70         \\
            OccFormer \cite{zhang2023occformer}                                  & 40.27    & 13.81        & 22.58        & 0.66        & 0.26        & 9.89         & 3.82         & 2.77        & 54.30        & 13.44        & 31.53        & 3.55        & 36.42 & 4.80        & 31.00        & 19.51 & 7.77         & 8.51        & 6.95         & 4.60         \\
            Symphonies \cite{jiang2023symphonize}    & 44.12   & 18.58 & \bf 30.02 & 1.85 & \bf 5.90 & \bf 25.07 & \bf 12.06 & \bf 8.20 & 54.94 & 13.83 & 32.76 & \bf 6.93 & 35.11        & \bf 8.58 & 38.33 & 11.52        & 14.01 & 9.57 & \bf 14.44 & \bf 11.28 \\
            CGFormer \cite{yu2024context}    & \bf 48.07 &\bf 20.05 & 29.85 &\bf 3.42 &3.96 &17.59 &6.79 &6.63 &\bf 63.85 &\bf 17.15 &\bf 40.72 &5.53 &\bf 42.73 &8.22 &\bf 38.80 &\bf 24.94 &\bf 16.24 &\bf 17.45 &10.18 &6.77  \\
            \bottomrule
        \end{tabular}
    \label{tab:kitti_360_test}
\end{table*}

SemanticKITTI \cite{semantickitti} is the first dataset with 3D occupancy labels for outdoor driving scenes. Occ3D-nuScenes \cite{occ3d} is the dataset used in the CVPR 2023 3D Occupancy Prediction Challenge \cite{CVPR_2023_3D_Occupancy_Prediction_Challenge}. These two datasets are currently the most popular. Therefore, we summarize the performance of various 3D occupancy methods that are trained and tested on these datasets, as reported in Tab. \ref{tab:semantikitt_performance} and \ref{tab:occ3d_performance}. Additionally, we evaluate the performance of 3D occupancy methods on the SSCBench-KITTI-360 dataset, as reported in Tab. \ref{tab:kitti_360_test}. These tables classify occupancy methods according to input modalities and supervised learning types, respectively. The best performances are highlighted in bold. Tab. \ref{tab:semantikitt_performance} and \ref{tab:kitti_360_test} utilize the IoU and mIoU metrics to evaluate the 3D geometric and 3D semantic occupancy perception capabilities. Tab. \ref{tab:occ3d_performance} adopts mIoU and mIoU$^{\ast}$ to assess 3D semantic occupancy perception. Unlike mIoU, the mIoU$^{\ast}$ metric excludes the 'others' and 'other flat' classes and is used by the self-supervised OccNeRF \cite{zhang2023occnerf}. For fairness, we compare the mIoU$^{\ast}$ of OccNeRF with other self-supervised occupancy methods. Notably, the OccScore metric is used in the CVPR 2024 Autonomous Grand Challenge \cite{CVPR_2024_Autonomous_Grand_Challenge}, but it has yet to become widely adopted. Thus, we do not summarize the occupancy performance with this metric. Below, we will compare perception accuracy from three aspects: overall comparison, modality comparison, and supervision comparison.

(1) \textit{Overall Comparison}. Tab. \ref{tab:semantikitt_performance} and \ref{tab:kitti_360_test} show that (i) the IoU scores of occupancy networks are less than 60\%, while the mIoU scores are less than 30\%. The IoU scores (indicating geometric perception, \textit{i.e.}, ignoring semantics) substantially surpass the mIoU scores. This is because predicting occupancy for some semantic categories is challenging, such as bicycles, motorcycles, persons, bicyclists, motorcyclists, poles, and traffic signs. Each of these classes has a small proportion (under 0.3\%) in the dataset, and their small sizes in shapes make them difficult to observe and detect. Therefore, if the IOU scores of these categories are low, they significantly impact the overall mIoU value. Because the mIOU calculation, which does not account for category frequency, divides the total IoU scores of all categories by the number of categories. (ii) A higher IoU does not guarantee a higher mIoU. One possible explanation is that the semantic perception capacity (reflected in mIoU) and the geometric perception capacity (reflected in IoU) of an occupancy network are distinct and not positively correlated.

Form Tab. \ref{tab:occ3d_performance}, it is evident that (i) the mIOU scores of occupancy networks are within 50\%, higher than the scores on SemanticKITTI and SSCBench-KITTI-360. For example, the mIOUs of TPVFormer \cite{huang2023tri} on SemanticKITTI and SSCBench-KITTI-360 are 11.26\% and 13.64\%, but it gets 27.83\% on Occ3D-nuScenes. OccFormer \cite{zhang2023occformer} and SurroundOcc \cite{surroundocc} have similar situations. We consider this might be due to the simpler task setting in Occ3D-nuScenes. On the one hand, Occ3D-nuScenes uses surrounding-view images as input, containing richer scene information compared to SemanticKITTI and SSCBench-KITTI-360, which only utilize monocular or binocular images. On the other hand, Occ3D-nuScenes only calculates mIOU for visible 3D voxels, whereas the other two datasets evaluate both visible and occluded areas, posing greater challenges. (ii) COTR \cite{ma2023cotr} has the best mIoU (46.21\%) and also achieves the highest scores in IoU across all categories on Occ3D-nuScenes.

(2) \textit{Modality Comparison}. The input data modality significantly influences 3D occupancy perception accuracy. Tab. \ref{tab:semantikitt_performance} and \ref{tab:kitti_360_test} report the performance of occupancy perception in different modalities. It can be seen that, due to the accurate depth information provided by LiDAR sensing, LiDAR-centric occupancy methods have more precise perception with higher IoU and mIoU scores. For example, on the SemanticKITTI dataset, S3CNet \cite{cheng2021s3cnet} has the top mIoU (29.53\%) and DIFs \cite{rist2021semantic} achieves the highest IoU (58.90\%); on the SSCBench-KITTI-360 dataset, S3CNet achieves the best IoU (53.58\%). However, we observe that the multi-modal approaches (\textit{e.g.}, OpenOccupancy \cite{openoccupancy} and Co-Occ \cite{pan2024co}) do not outperform single-modal (\textit{i.e.}, LiDAR-centric or vision-centric) methods, indicating that they have not fully leveraged the benefits of multi-modal fusion and the richness of input data. Therefore, there is considerable potential for further improvement in multi-modal occupancy perception. Moreover, vision-centric occupancy perception has advanced rapidly in recent years. On the SemanticKITTI dataset, the state-of-the-art vision-centric occupancy methods still lag behind LiDAR-centric methods in terms of IoU and mIoU. But notably, the mIoU of the vision-centric CGFormer \cite{yu2024context} has surpassed that of LiDAR-centric methods on the SSCBench-KITTI-360 dataset.

(3) \textit{Supervision Comparison}. The 'Sup.' column of Tab. \ref{tab:occ3d_performance} outlines supervised learning types used for training occupancy networks. Training with strong supervision, which directly employs 3D occupancy labels, is the most prevalent type. Tab. \ref{tab:occ3d_performance} shows that occupancy networks based on strongly-supervised learning achieve impressive performance. The mIoU scores of FastOcc \cite{hou2024fastocc}, FB-Occ \cite{li2023fb}, PanoOcc \cite{wang2023panoocc}, and COTR \cite{ma2023cotr} are significantly higher (12.42\%-38.24\% increased mIoU) than those of weakly-supervised or self-supervised methods. This is because occupancy labels provided by the dataset are carefully annotated with high accuracy, and can impose strong constraints on network training. However, annotating these dense occupancy labels is time-consuming and laborious. It is necessary to explore network training based on weak or self supervision to reduce reliance on occupancy labels. Vampire \cite{xu2024regulating} is the best-performing method based on weakly-supervised learning, achieving a mIoU score of 28.33\%. It demonstrates that semantic LiDAR point clouds can supervise the training of 3D occupancy networks. However, the collection and annotation of semantic LiDAR point clouds are expensive. SelfOcc \cite{huang2023selfocc} and OccNeRF \cite{zhang2023occnerf} are two representative occupancy works based on self-supervised learning. They utilize volume rendering and photometric consistency to acquire self-supervised signals, proving that a network can learn 3D occupancy perception without any labels. However, their performance remains limited, with SelfOcc achieving an mIoU of 7.97\% and OccNeRF an mIoU$^{\ast}$ of 10.81\%.

\subsubsection{Inference Speed}

\begin{table}[t] 
    \footnotesize
    \setlength{\tabcolsep}{1.3mm}
    \centering
    \caption{ \textbf{Inference speed analysis of 3D occupancy perception on the Occ3D-nuScenes \cite{occ3d} dataset.} $\dagger$ indicates data from  SparseOcc \cite{liu2024fully}. $\ddagger$ means data from FastOcc \cite{hou2024fastocc}. R-50 represents ResNet50 \cite{he2016deep}. TRT denotes acceleration using the TensorRT SDK \cite{vanholder2016efficient}.
    }
    \begin{tabular}{l|c|c|c|c|c}
    \toprule[1.5pt]
    Method & GPU &  Input Size & Backbone & mIoU(\%) & FPS(Hz) \\  
    \midrule
    BEVDet$\dagger$ \cite{huang2021bevdet}         & A100 &  704$\times$ 256 & R-50 & 36.10 &2.6 \\
    BEVFormer$\dagger$ \cite{li2022bevformer}         & A100 &  1600$\times$900 & R-101 & 39.30 &3.0 \\
    FB-Occ$\dagger$ \cite{li2023fb}         & A100 &  704$\times$256 & R-50 & 10.30 &10.3 \\
     SparseOcc \cite{liu2024fully}         & A100 &  704$\times$256 & R-50 & 30.90 &12.5 \\
    
     \midrule
     SurroundOcc$\ddagger$ \cite{surroundocc}   & V100  & 1600$\times$640& R-101 &37.18 &2.8 \\
    FastOcc \cite{hou2024fastocc}  & V100 &  1600$\times$640 & R-101 &40.75 &4.5 \\
    FastOcc(TRT) \cite{hou2024fastocc} & V100 & 1600$\times$640& R-101&40.75 &12.8\\
    \bottomrule[1.5pt]
    \end{tabular}

    \label{tab:inference_speed}
\end{table}

Recent studies on 3D occupancy perception \cite{hou2024fastocc,liu2024fully} have begun to consider not only perception accuracy but also its inference speed. According to the data provided by FastOcc \cite{hou2024fastocc} and  SparseOcc \cite{liu2024fully}, we sort out the inference speeds of 3D occupancy methods, and also report their running platforms, input image sizes, backbone architectures, and occupancy accuracy on the Occ3D-nuScenes dataset, as depicted in Tab. \ref{tab:inference_speed}.

A practical occupancy method should have high accuracy (mIoU) and fast inference speed (FPS). From Tab. \ref{tab:inference_speed}, FastOcc achieves a high mIoU (40.75\%), comparable to the mIOU of BEVFomer. Notably, FastOcc has a higher FPS value on a lower-performance GPU platform than BEVFomer. Furthermore, after being accelerated by TensorRT \cite{vanholder2016efficient}, the inference speed of FastOcc reaches 12.8Hz.

\section{Challenges and Opportunities}
\label{sec:Challenges and Opportunities}
In this section, we explore the challenges and opportunities of 3D occupancy perception in autonomous driving. Occupancy, as a geometric and semantic representation of the 3D world, can facilitate various autonomous driving tasks. We discuss both existing and prospective applications of 3D occupancy, demonstrating its potential in the field of autonomous driving. Furthermore, we discuss the deployment efficiency of occupancy perception on edge devices, the necessity for robustness in complex real-world driving environments, and the path toward generalized 3D occupancy perception.

\subsection{Occupancy-based Applications in Autonomous Driving }
3D occupancy perception enables a comprehensive understanding of the 3D world and supports various tasks in autonomous driving. Existing occupancy-based applications include segmentation, detection, dynamic perception, world models, and autonomous driving algorithm frameworks. (1) Segmentation: Semantic occupancy perception can essentially be regarded as a 3D semantic segmentation task. (2) Detection: OccupancyM3D \cite{peng2023learning} and SOGDet \cite{zhou2024sogdet} are two occupancy-based works that implement 3D object detection. OccupancyM3D first learns occupancy to enhance 3D features, which are then used for 3D detection. SOGDet develops two concurrent tasks: semantic occupancy prediction and 3D object detection, training these tasks simultaneously for mutual enhancement. 
(3) Dynamic perception: Its goal is to capture dynamic objects and their motion in the surrounding environment, in the form of predicting occupancy flows for dynamic objects. Strongly-supervised Cam4DOcc \cite{cam4docc} and self-supervised LOF \cite{liu2024let} have demonstrated potential in occupancy flow prediction.
(4) World model: It simulates and forecasts the future state of the surrounding environment by observing current and historical data \cite{lecun2022path}. Pioneering works, according to input observation data, can be divided into semantic occupancy sequence-based world models (\textit{e.g.}, OccWorld \cite{zheng2023occworld} and OccSora \cite{wang2024occsora}), point cloud sequence-based world models (\textit{e.g.}, SCSF \cite{wang2024semantic}, UnO \cite{agro2024uno}, PCF \cite{khurana2023point}), and multi-camera image sequence-based world models (\textit{e.g.}, DriveWorld \cite{min2024driveworld} and Cam4DOcc \cite{cam4docc}). However, these works still perform poorly in high-quality long-term forecasting.
(5) Autonomous driving algorithm framework: It integrates different sensor inputs into a unified occupancy representation, then applies the occupancy representation to a wide span of driving tasks, such as 3D object detection, online mapping, multi-object tracking, motion prediction, occupancy prediction, and motion planning. Related works include OccNet \cite{openocc}, DriveWorld \cite{min2024driveworld}, and UniScene \cite{min2024multi}.

However, existing occupancy-based applications primarily focus on the perception level, and less on the decision-making level. Given that 3D occupancy is more consistent with the 3D physical world than other perception manners (\textit{e.g.,} bird's-eye view perception and perspective-view perception), we believe that 3D occupancy holds opportunities for broader applications in autonomous driving. At the perception level, it could improve the accuracy of existing place recognition \cite{xu2023novel,xu2024c2l}, pedestrian detection \cite{jain2023multimodal,guan2019fusion}, accident prediction \cite{wang2024deepaccident}, and lane line segmentation \cite{zou2022novel}. At the decision-making level, it could help safer driving decisions \cite{xu2020explainable} and navigation \cite{zhuang2023multi,li2023multi}, and provide 3D explainability for driving behaviors.

\subsection{Deployment Efficiency}
For complex 3D scenes, large amounts of point cloud data or multi-view visual information always need to be processed and analyzed to extract and update occupancy state information. To achieve real-time performance for the autonomous driving application, solutions commonly need to be computationally complete in a limited amount of time and need to have efficient data structures and algorithm designs. In general, deploying deep learning algorithms on target edge devices is not an easy task. 

Currently, some real-time and deployment-friendly efforts on occupancy tasks have been attempted. For instance, Hou \textit{et al.} \cite{hou2024fastocc} proposed a solution, FastOcc, to accelerate prediction inference speed by adjusting the input resolution, view transformation module, and prediction head. Zhang \textit{et al.} \cite{zhang2024bdc} further lightweighted FlashOcc by decomposing its occupancy network and binarizing it with binarized convolutions. Liu \textit{et al.} \cite{liu2024fully} proposed SparseOcc, a sparse occupancy network without any dense 3-D features, to minimize computational costs using sparse convolution layers and mask-guided sparse sampling. Tang \textit{et al.} \cite{tang2024sparseocc} proposed to adopt sparse latent representations and sparse interpolation operations to avoid information loss and reduce computational complexity. Additionally, Huang \textit{et al.} recently proposed GaussianFormer \cite{huang2024gaussianformer}, which utilizes a series of 3D Gaussians to represent sparse interest regions in space. GaussianFormer optimizes the geometric and semantic properties of the 3D Gaussians, corresponding to the semantic occupancy of the interest regions. It achieves comparable accuracy to state-of-the-art methods using only 17.8\%-24.8\% of their memory consumption. However, the above-mentioned approaches are still some way from practical deployment in autonomous driving systems. A deployment-efficient occupancy method requires superiority in real-time processing, lightweight design, and accuracy simultaneously.

\subsection{Robust 3D Occupancy Perception}
In dynamic and unpredictable real-world driving environments, the perception robustness is crucial to autonomous vehicle safety. State-of-the-art 3D occupancy models may be vulnerable to out-of-distribution scenes and data, such as changes in lighting and weather, which would introduce visual biases, and input image blurring, which is caused by vehicle movement. Moreover, sensor malfunctions (\textit{e.g.}, loss of frames and camera views) are common \cite{huang2023multi}. In light of these challenges, studying robust 3D occupancy perception is valuable.

However, research on robust 3D occupancy is limited, primarily due to the scarcity of datasets. Recently, the ICRA 2024 RoboDrive Challenge \cite{ICRA_2024_RoboDrive_Challenge} provides imperfect scenarios for studying robust 3D occupancy perception.

In terms of network architecture and scene representation, we consider that related works on robust BEV perception \cite{liu2023bevfusion,liang2022bevfusion,xie2023robobev,chen2024m,GeokernelTransformer,kim2023crn} could inspire developing robust occupancy perception. M-BEV \cite{chen2024m} proposes a masked view reconstruction module to enhance robustness under various missing camera cases. GKT \cite{GeokernelTransformer} employs coarse projection to achieve robust BEV representation. In terms of sensor modality, radar can penetrate small particles such as raindrops, fog, and snowflakes in adverse weather conditions, thus providing reliable detection capability. Radar-centric RadarOcc \cite{ding2024radarocc} achieves robust occupancy perception with imaging radar, which not only inherits the robustness of mmWave radar in all lighting and weather conditions, but also has higher vertical resolution than mmWave radar. RadarOcc has demonstrated more accurate 3D occupancy prediction than LiDAR-centric and vision-centric methods in adverse weather. Besides, in most damage scenarios involving natural factors, multi-modal models \cite{liu2023bevfusion,liang2022bevfusion,kim2023crn} usually outperform single-modal models, benefiting from the complementary nature of multi-modal inputs. In terms of training strategies, Robo3D \cite{kong2023robo3d} distills knowledge from a teacher model with complete point clouds to a student model with imperfect input, enhancing the student model's robustness. Therefore, based on these works, approaches to robust 3D occupancy perception could include, but are not limited to, robust scene representation, multiple modalities, network design, and learning strategies.

\subsection{Generalized 3D Occupancy Perception}
Although more accurate 3D labels mean higher occupancy prediction performance \cite{kalble2024accurate}, 3D labels are costly and large-scale 3D annotations for the real world are impractical. The generalization capabilities of existing networks trained on limited 3D-labeled datasets have not been extensively studied. To get rid of dependence on 3D labels, self-supervised learning represents a potential pathway toward generalized 3D occupancy perception. It learns occupancy perception from a broad range of unlabelled images. However, the performance of current self-supervised occupancy perception \cite{gan2023simple,zhang2023occnerf,huang2023selfocc,han2024boosting} is poor. On the Occ3D-nuScene dataset (see Tab. \ref{tab:occ3d_performance}), the top accuracy of self-supervised methods is inferior to that of strongly-supervised methods by a large margin. Moreover, current self-supervised methods require training and evaluation with more data. Thus, enhancing self-supervised generalized 3D occupancy perception is an important future research direction.

Furthermore, current 3D occupancy perception can only recognize a set of predefined object categories, which limits its generalizability and practicality. Recent advances in large language models (LLMs) \cite{chung2024scaling,zheng2024judging,Chatgpt,touvron2023llama} and large visual-language models (LVLMs) \cite{zhu2023minigpt,liu2024visual,achiam2023gpt,dai2024instructblip,zhou2022extract} demonstrate a promising ability for reasoning and visual understanding. Integrating these pre-trained large models has been proven to enhance generalization for perception \cite{vobecky2024pop}. POP-3D \cite{vobecky2024pop} leverages a powerful pre-trained visual-language model \cite{zhou2022extract} to train its network and achieves open-vocabulary 3D occupancy perception. Therefore, we consider that employing LLMs and LVLMs is a challenge and opportunity for achieving generalized 3D occupancy perception.

\section{Conclusion}
\label{sec:Conclusions}
This paper provided a comprehensive survey of 3D occupancy perception in autonomous driving in recent years. We reviewed and discussed in detail the state-of-the-art LiDAR-centric, vision-centric, and multi-modal perception solutions and highlighted information fusion
techniques for this field. To facilitate further research, detailed performance comparisons of existing occupancy methods are provided. Finally, we described some open challenges that could inspire future research directions in the coming years. We hope that this survey can benefit the community, support further development in autonomous driving, and help inexpert readers navigate the field.

\section*{Acknowledgment}
The research work was conducted in the JC STEM Lab of Machine Learning and Computer Vision funded by The Hong Kong Jockey Club Charities Trust and was partially supported by the Research Grants Council of the Hong Kong SAR, China (Project No. PolyU 15215824).

\bibliography{IF}

\clearpage

\end{sloppypar}
\end{document}